\documentclass[
10pt, 
a4paper, 
oneside, 
headinclude,footinclude, 
BCOR5mm, 
]{scrartcl}

\usepackage{float}
\usepackage{multirow}
\usepackage{orcidlink}
\usepackage{xcolor} 
\definecolor{Maroon}{RGB}{128, 0, 0} 
\definecolor{RoyalBlue}{cmyk}{1, 0.50, 0, 0} 

\usepackage[
nochapters, 
beramono, 
eulermath,
pdfspacing, 
dottedtoc 
]{classicthesis} 

\usepackage{arsclassica} 

\usepackage[T1]{fontenc} 

\usepackage[utf8]{inputenc} 

\usepackage{graphicx} 
\graphicspath{{Figures/}} 

\usepackage{enumitem} 

\usepackage{lipsum} 

\usepackage{subfig} 

\usepackage{amsmath,amssymb,amsthm} 

\usepackage{varioref} 

\theoremstyle{definition} 

\theoremstyle{plain} 

\theoremstyle{remark} 

\hypersetup{
colorlinks=true, breaklinks=true, bookmarks=true,bookmarksnumbered,
urlcolor=webbrown, linkcolor=webbrown, citecolor=webgreen, 
pdftitle={}, 
pdfauthor={\textcopyright}, 
pdfsubject={}, 
pdfkeywords={}, 
pdfcreator={pdfLaTeX}, 
pdfproducer={LaTeX with hyperref and ClassicThesis} 
} 

\title{\normalfont\spacedallcaps{Comparing Deep Learning models for rice mapping in Bhutan using high resolution satellite imagery}}

\author{\spacedlowsmallcaps{Biplov Bhandari* \orcidlink{0000-0001-6169-8236} \textsuperscript{1, 2} \& Timothy Mayer \orcidlink{0000-0001-9489-9392}\textsuperscript{1, 2}}} 

\date{} 

\begin{document}


\renewcommand{\sectionmark}[1]{\markright{\spacedlowsmallcaps{#1}}} 
\lehead{\mbox{\llap{\small\thepage\kern1em\color{halfgray} \vline}\color{halfgray}\hspace{0.5em}\rightmark\hfil}} 

\pagestyle{scrheadings} 


\maketitle 






\section*{Abstract} 

Keywords: Bhutan; Crop type mapping; Rice Mapping; Deep Learning; Planet

Crop type and crop extent are critical information that help policy makers take informed decision on food security. As the economic growth of Bhutan has increased at an annual rate of 7.5\% over the last three decades, it is very important to understand how this economic and population growth affects and impacts on food security. Through various policies and implementation, the Bhutanese government is promoting a number of drought-resilient, high yielding, and disease-resistant crop varieties. Simultaneously the Bhutanese government is increasing its utilization of technological approaches such as including Remote Sensing-based knowledge and data products into their decision-making process. This study focuses on Paro, one of the top rice yielding district in Bhutan and employs publicly available Norway's International Climate and Forest Initiative (NICFI) high resolution satellite imagery from Planet. Two Deep Learning approaches,  point-based (DNN) and patch based (U-Net), models were used in conjunction with cloud-computing platforms. Three different models per Deep Learning approaches (DDN and U-Net) were trained: 1) RGBN channels from Planet,  2) RGBN and elevation data (RGBNE), 3) RGBN and Sentinel-1 data (RGBNS), and RGBN with elevation and Sentinel-1 data (RGBNES). From this comprehensive analysis the U-Net displayed higher performance metrics across both model training and model validation efforts. Among the U-Net model sets, the RGBN, RGBNE, RGBNS, and RGBNES models had an F1-score of 0.8546, 0.8563, 0.8467, and 0.8500 respectively. An additional  separate independent model evaluation was performed and found a high level of performance variation  across all the metrics (precision, recall, and F1-score) underscoring the need for practitioners to employ independent validation. For this independent model evaluation, the U-Net RGBN, RGBNE, RGBNES, and RGBN models displayed the F1-scores of 0.5935, 0.6154, 0.5882, and 0.6582, suggesting U-Net RGBNES as the best model across the comparison. The study demonstrates that the  Deep Learning approaches are able to predict at the crop level, mapping rice. Also Deep Learning methods can be used in combination with the survey based approaches currently utilized by the Department of Agriculture (DoA) in Bhutan. Further this study  successfully demonstrated the usage of regional land cover products such as SERVIR's Regional Land Cover Monitoring System (RLCMS) as a weak label approach to capture different strata addressing the class imbalance problem and improving the sampling design for Deep Learning application. Finally through preliminary model testing and comparisons outlined it was demonstrated that using additional features such as NDVI, EVI, and NDWI did not drastically improve model performance.




\let\thefootnote\relax\footnotetext{\textsuperscript{1} \textit{Earth System Science Center, The University of Alabama in Huntsville, 320 Sparkman Drive, Huntsville, AL 35805, USA}}

\let\thefootnote\relax\footnotetext{\textsuperscript{2} \textit{SERVIR Science Coordination Office, NASA Marshall Space Flight Center, 320 Sparkman Drive, Huntsville, AL 35805, USA}}


\newpage 


\section{Introduction}

With population trends continuing to increase, the need for food security at a national level is essential \cite{food2017future}. 
In the past decades, south Asia and southeast Asia has undergone rapid economic growth and extensive structural changes such as urbanization \cite{liu2020agricultural}. Due to this vast changes, it is critical to effectively manage, understand and promote improvised farm and supply management approaches. For this, understanding on the type, extent, cycle, and duration of the stable crop both spatially and temporally is essential. In particular, the information on crop type and extent can also be used in variety of downstream applications including crop yield estimation, understanding the effect of natural hazards including flood and droughts in agricultural applications, and hydrological models.

There are a wide range of Remote Sensing based approaches for the identification and delineation of crop type and crop extent \cite{weiss2020remote}. These studies use an array of satellite-based sensors and data for mapping the type and extent of crops. For example, Tariq et al 2023 \cite{tariq2023mapping} used optical images from Sentinel-2 and Landsat-8 Normalized Difference Vegetation Index (NDVI) to map the type and pattern of specific crops. Other studies have used threshold based approaches \cite{lobell2004cropland, pan2012winter}, and indices-based approaches including NDVI, Enhanced Vegetation Index (EVI) and Land Surface Water Index (LSWI) \cite{XIAO2005480, XIAO200695}. In several instances, the phenology based methods \cite{dong2016mapping, mishra2023high} have been employed with optical or radar data \cite{park2018classification}. Different studies have employed diverse methodologies for example statistical \cite{ma2020unsupervised}, clustering \cite{mayaux2004new}, fusion techniques \cite{gao2006blending, zhu2010enhanced, gevaert2015comparison}, Machine Learning such as Support Vector Machines \cite{carrao2008contribution, zhang2009mapping}, Decision trees \cite{gebhardt2014mad, mayer2023employing, o2020improved, yu2023ricemapengine, clark2012land, singha2019high, lasko2018mapping}, and Deep Learning \cite{poortinga2021mapping, mayer2021deep, parekh2021automatic, lv2020delineation}. When selecting these methods, consideration such as the availability of data, type of data, statistical distribution of classes, target accuracy, scalability, and transferability should be considered as these often exhibit direct trade-offs \cite{gomez2016optical}.

Rice is a primary food staple in Bhutan feeding about half of the total population \cite{tashi2018mapping} and rice holds cultural significance acting as an essential driver of national food security within the country \cite{tshewang2016weed}. Bhutan, over the past three decades, displayed a 7.5\% annual average economic growth rate \cite{WB_Bhutan}. Simultaneously Bhutan is taking measures to eradicate poverty, with a reduction from 31\% in 2003 to 8\% in 2017 \cite{world2019bhutan}. With such advancements in economic growth, food security is essential for sustainable development, thus identification and delineation of crop type and crop extent at a fine spatial resolution is extremely valuable for decision-makers to make informed decisions.

While, there are some global or regional landcover data sets that have cropland layers, e.g., the MCD12Q1 from the Moderate Resolution Imaging Spectroradiometer (MODIS) \cite{FRIEDL2010168}, the Copernicus global land cover layers \cite{buchhorn2020copernicus}, Dynamic World from the Google \cite{brown2022dynamic}, and the Regional Land Cover Monitoring System (RLCMS) from the SERVIR \cite{SAAH2020101979, uddin2021regional}. In general these broader map products include a crop category, but do not contain the crop type information such as  rice paddy. While other studies have looked into mapping rice extent in South Asia with several studies extending to Bhutan \cite{XIAO200695, gumma2011mapping}, however these mostly uses moderate scale MODIS (500m). In addition, Bhutan is a particularly challenging country to explore the incorporation of Remote Sensing approaches for monitoring because of: 1) its complex terrain; 2) cloud cover and limited quantity of Earth observations data; 3) regionally specific cultivation practices relaying on rain-fed irrigation, 4) small farm holdings ($\sim$ 1.2 hectares \cite{WB_2017}) in most cultivation areas, 5) scarcity of ground truth labels, and 6) a wide variety of agricultural commodities beyond rice including maize, tree crop commodities etc.

Both the Bhutanese government and general public of Bhutan has expressed a strong desire to embrace and utilize technological approaches such as Remote Sensing based monitoring systems into their decision-making \cite{gurung2023identification, dorji2023agricultural}. Recent examples include the collaborative efforts through the Advancing Science, Technology, Engineering, and Mathematics in Bhutan through Increased Earth Observation Capacity initiative \cite{mccartney2021advancing} that included developing applications for water resources management, ecological forecasting, and agriculture related applications. One of the outcome of this engagement is the co-development of the Agricultural Classification and Estimation Service (ACES) platform \cite{mayer2023employing}. A culmination of this work provided a cloud-based processing pipeline for producing yearly rice maps using Random Forest \cite{breiman2001random} in Google Earth Engine (GEE) \cite{gorelick2017google} and a \href{https://github.com/SERVIR/ag-classification-estimation?tab=readme-ov-file}{GEE-based web interface} for viewing the rice maps and other relevant meteorological data. However, moderate spatial resolution and strong cloud prevalence some of the limitations of the initial ACES platform. Therefore reliable, accurate, and high resolution crop type maps in Bhutan remains limited and accurate crop type maps are needed. This study aims to fill this gap by using remote sensing-based data along with the farmer knowledge at the field level to produce accurate high resolution crop type and crop extent maps by deploying a Deep Learning (DL) modeling techniques.

The objectives of this study were to: 1) produce 10 m scale crop type maps leveraging high resolution satellite imagery for Paro, one of the most productive rice growing Dzongkhag (districts) in Bhutan, 2) assess the importance of indices as feature engineering and employing indices as features in crop mapping; 3) explore the application of two different deep learning model architectures (Deep Neural Networks (DNN) and U-Net) at both a pixel and patch level. The results will help policy makers in Bhutan to make informed decision through leveraging the high resolution crop type maps. Additionally this study aims to  build capacity with partners for using DL workflows. As existing area estimation methods are based off of survey-based method \cite{NSB_Ag_2019, NSB_Ag_2020, NSB_Ag_2021}, if applied, the developed Remote Sensing based method can also help with managing on-the ground resources. This research informs part of SERVIR's work on STEM engagement and service co-development by SERVIR and the Bhutanese partners at the Department of Agriculture \cite{Figueredo:2009dg}.

\section{Study Area}
\subsection {Agriculture practices in Bhutan}
The Kingdom of Bhutan is a mountainous landlocked country between India and China with complex terrain. Bhutan has an area of 38,394 km\textsuperscript{2} \cite{gilani2015decadal, NSB2023} with a projected population of 770,276 in 2023 \cite{NSB2023}. Forest is the major land cover in the country occupying 69\% \cite{NSB2023, DoSAM2023} excluding Alpine Scrubs and Shrubs which occupies 8.89\% and 4.11\% respectively \cite{DoSAM2023}. Additionally, agricultural land use covers 2.96\% equivalent to about ~281,000 acres. The land cover land use changes in the country can be attributed to climate change, natural disasters, deforestation, urbanization, land degradation, and policy changes \cite{NSB2023, gyeltshen2019integrate}.

Agriculture in Bhutan is mostly small scale, and depends on the monsoon rain. Rice and maize are the major cereal crops in the country making of more than 85\% of the total cereal crop production \cite{NSB_Ag_2021}. Farming is mostly subsistence basis with little market surplus. However, in the recent years, with the increase of the transportation network and market connectivity, farmers are switching from subsistence based farming to commercial farming \cite{gyeltshen2019integrate, Agriculture2017}. In addition, in recent years, Department of Agriculture (DoA) has introduced a number of drought-resilient, high yielding, and disease-resistant rice, maize and wheat crop varieties across various agro-ecological zones \cite{chhogyel2018climate}.

\begin{figure}[hbt]
\centering 
\includegraphics[width=\columnwidth]{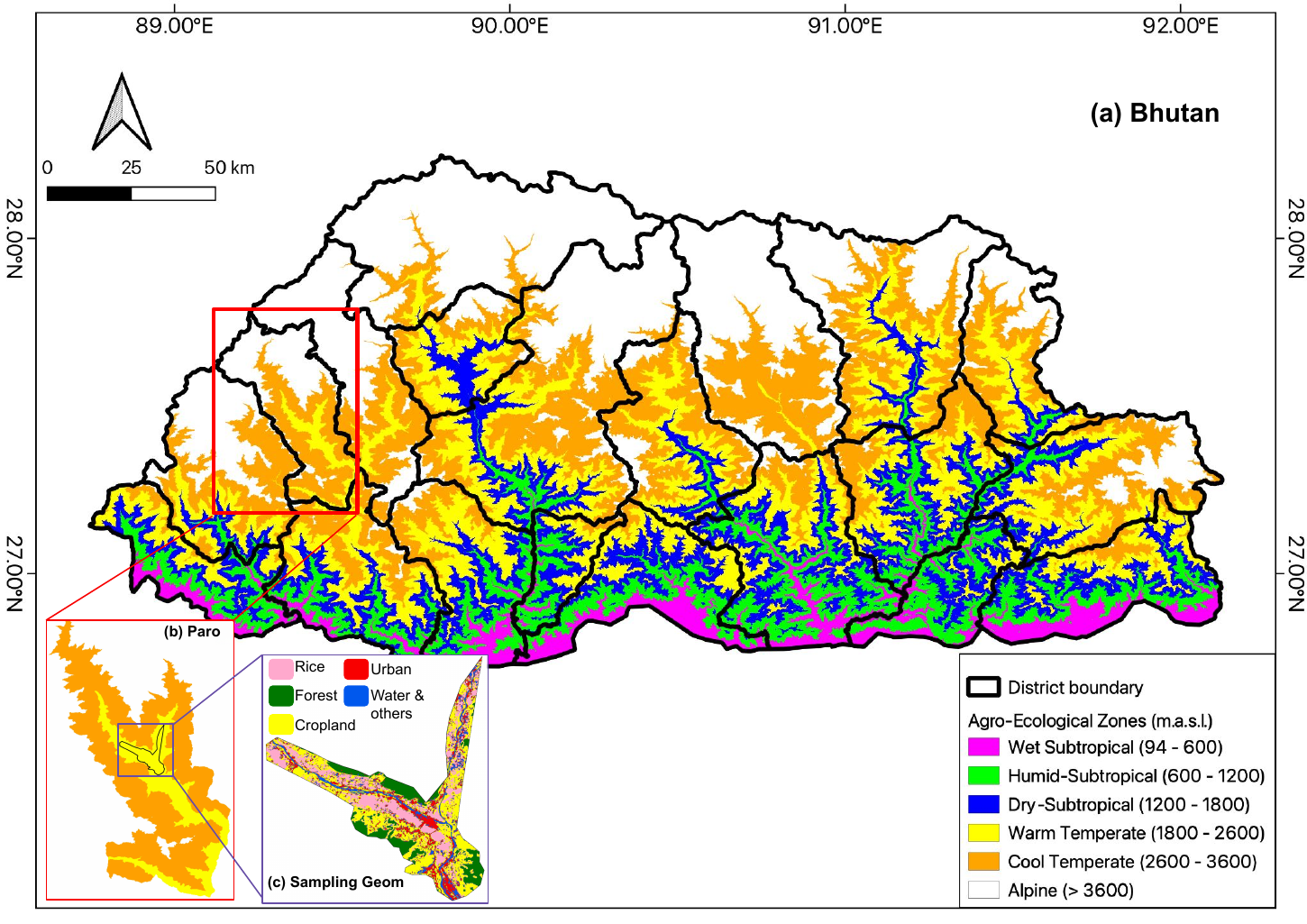}
\caption[(a) Map of Bhutan. The country is divided into six different Agro-Ecological Zones (AEZs) based on the altitude: Wet-Subtropical, Humid-Subtropical, Dry-Subtropical, Warm Temperate, Cool Temperate and Alpine. (b) Study Area Map of Paro Dzongkhag. Rice is normally grown on Warm Temperature AEZ of Paro (between 1900 to 2600 m.s.l. This study focuses on this rice growing elevation range. (c) Sampling geometry for generating training data for DL algorithms.]{(a) Map of Bhutan. The country is divided into six different Agro-Ecological Zones (AEZs) based on the altitude: Wet-Subtropical, Humid-Subtropical, Dry-Subtropical, Warm Temperate, Cool Temperate and Alpine. (b) Study Area Map of Paro Dzongkhag. Rice is normally grown on Warm Temperature AEZ (between 1900 to 2600 m.s.l. This study focuses on this rice growing elevation range. (c) Sampling geometry for generating training data for DL algorithms.}
\label{fig:study_area_map}
\end{figure}

The country is divided into six different Agro-Ecological Zones (AEZs) based on the altitude: Wet-Subtropical (94-600 meters above sea level (m.a.s.l.), Humid-Subtropical (600-1200 m.a.s.l.), Dry-Subtropical (1200-1800 m.a.s.l.), Warm Temperate (1800-2600 m.a.s.l.), cool temperate (2600-3600 m.a.s.l.) and alpine (> 3600 m.a.s.l.) \cite{Agriculture2017, gyeltshen2019integrate}. These different zone have distinct vegetation and unique agriculture practices. The three subtropical zones (wet, humid, and dry) are located in the Southern portion of Bhutan and are located at the foothills of the Himalayan mountains. These regions receive heavy precipitation and experience high humidity. The major crop grown in these zones are rice (paddy), maize, millet, legume/pulses, and citrus. The temperate zones are usually characterized by hot summer and cool winter. The warm temperate AEZ receives similar rain as to the dry subtropical zone, but this AEZ is usually less humid. The warm temperate AEZ is suitable for rice, wheat, mustard, fruits, and vegetables. In the cool temperate AEZ, vegetables such as potato and buckwheat are grown. In addition, in this AEZ livestock are also prevalent. Finally, the alpine zone consists of mostly permanent snows and glaciers with very cold winters and cool summers \cite{gyeltshen2019integrate}. Figure ~\ref{fig:study_area_map} displays the AEZ of Bhutan along with the study area.

\subsection {Rice practices in Paro}
This study focuses on the Paro district of Bhutan. Paro is located in the north-western region of Bhutan and extends from the foothills to high-mountainous region of Bhutan. Paro has an area of approximately 1285.09 km\textsuperscript{2} displayed in figure ~\ref{fig:study_area_map}. Paro is well known as the widest valley in the country with extensive fertile rice fields located along the Pa Chhu, the local name for the Paro River \cite{tashi2018mapping}. The elevation of Paro ranges from the lowest of 1946 m to the highest of 5656 m with an average elevation of 3521 m. Paro is one of the top three rice growing Dzongkhag in Bhutan.

The dominant land cover in the area is forest ($\sim$ 47.14 \%) while agriculture makes about 5.35\% \cite{DoSAM2023}. Rice is usually planted between 1900 to 2600 m.a.s.l. which is the Warm Temperate AEZ. This study focuses on this rice growing elevation range. The majority of rice cultivated here is considered irrigated rice and farmers plant within both the valley floors and valley walls (terraced type) small fields, with majority being the valley floors. While the upland rice (Kam Bja/Pang bara) also exists in Paro, it is cultivated in much smaller quantities. In 2021, farmers in Paro sowed 2,064.62 acre of irrigated paddy, while only sowed 2.28 acre of upland paddy \cite{NSB_Ag_2021}. This diversity of cultivation and farming practices makes Paro a good representative district to develop the DL methods for the country.  According to the Ministry of Agriculture and Forests (MoAF) report, about 8.30\%, 9.19\%, and 7.95\% of all the rice was irrigated in Paro with total production of about 12.95\%, 15.03\%, and 12.91\% in year 2019, 2020, and 2021 respectively making it one of the top rice growing districts \cite{NSB_Ag_2019, NSB_Ag_2020, NSB_Ag_2021}.


Rice is cultivated every year from June to September in the Paro Dzongkhag \cite{Namgay2021}. Other winter vegetables and crops like wheat, buckwheat, and millet are also planted, but are often in smaller production and total area. Local rice is broadly classified as maap (red rice) or kaap (white rice); the maap which is predominately popular in higher altitude are usually planted in Paro, while the lower altitude region usually grow kaap. Some popular variety of rice in Paro consists of Kochum, Rey Kaap, Naam, Hasey, Kam Bja, among others \cite{jena2012advances}. Rice cultivation in Bhutan consists of four major growth stages: (1) Germination including seedling and its establishment; (2) Tillering; (3) Anthesis which is the flowering and maturation stage; (4) and Ripening stage \cite{Namgay2021}. Since the high-altitude environment has a temperature pattern and cold air; the low temperature is a problem in the early growth stage and sometimes in the reproductive and ripening stages \cite{jena2012advances}.


\section{Data and Methods}

\subsection{Planet's satellite imagery}
For the optical remote sensing data set, Planetscope basemap was obtained as part of the Norway's International Climate and Forest Initiative (NICFI) data program \cite{planet_nicfi_data}. The Planet surface reflectance mosaic data are provided in an analysis ready format optimized for scientific and quantitative analysis \cite{planet_nicfi_data}. The monthly data are provided for the Blue (B) (455–515 nm), Green (G) (500–590 nm), Red (R) (590–670 nm), and NIR (N) (780–860 nm) (referred to as RGBN in subsequent analysis) spectral resolution at 4.77 m spatial resolution, applying atmospheric and cloud corrections, thus enhancing the quality of the mosaic \cite{planet_nicfi_data, lemajic2018new}.

\begin{figure}[hbt]
\centering 
\includegraphics[width=\columnwidth]{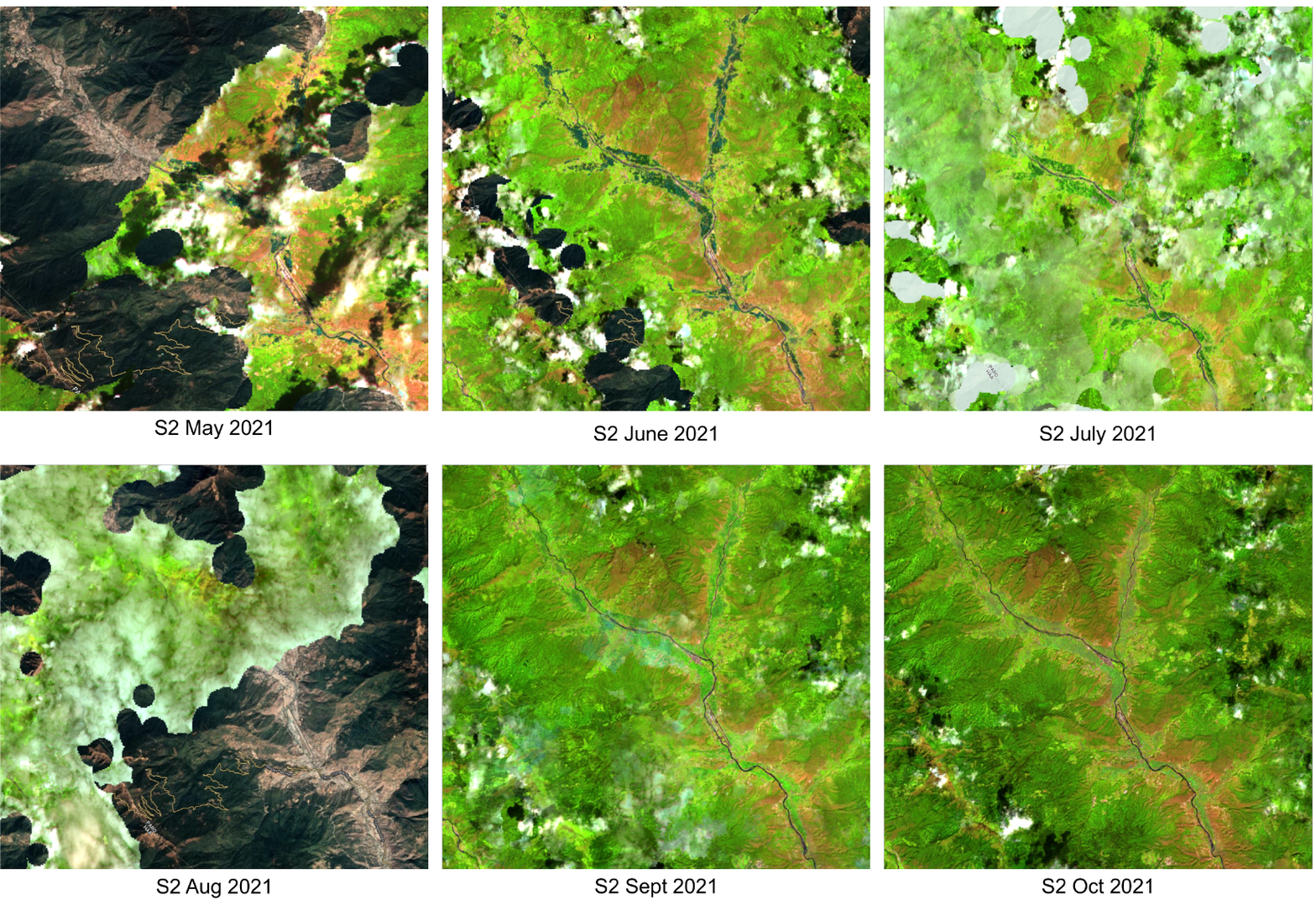} 
\caption[Sentinel-2 (S2) composites per month for the study year 2021 over the Paro district of Bhutan. Due to monsoon based farming, the data scarcity is an issue when using S2 images.]{Sentinel-2 (S2) composites per month for the study year 2021 over the Paro district of Bhutan. Due to monsoon based farming, the data scarcity is an issue when using S2 images.}
\label{fig:s2_composite_per_month} 
\end{figure}

The Planet mosaic data set were preferred over the Sentinel-2 (S2) or Landsat Multispectral Instrument (MSI) because of the cloud cover persistent in the region. The composite coverage of the S2 images over Paro for the year 2021 from May to October are shown in Figure \ref{fig:s2_composite_per_month}. The Planet monthly mosaic were obtained from the GEE.

The composite images of pre-growing season and growing season (June to September) were used. Since the valley of the Paro is surrounded by the evergreen forest, the different composite makes it easier to differentiate between the classes. Additionally, derived indices from the RGBN bands were constructed, namely NDVI, EVI, Normalized Difference Water Index (NDWI), Soil-adjusted Vegetation Index (SAVI), Modified Soil-adjusted Vegetation Index (MSAVI), Modified Triangular Vegetation Index (MTVI), Visual Atmosphere Resistance Index (VARI), and Triangular Greenness Index (TGI) \cite{planet_indices}.

The formula to calculate each of them are as:

\begin{equation}
NDVI =\frac{N - R}{N + R}
\label{eq:ndvi}
\end{equation}

\begin{equation}
EVI =2.5*\frac{N - R}{N + 6*R -7.5*B + 1}
\label{eq:evi}
\end{equation}

\begin{equation}
NDWI =\frac{G - N}{G + N}
\label{eq:ndwi}
\end{equation}

\begin{equation}
SAVI = \frac{N - R}{N + R + 0.5} * 1.5
\label{eq:savi}
\end{equation}

\begin{equation}
MSAVI =\frac{2 * N + 1 - \sqrt{(2 * N + 1)^2 - 8* (N - R)}}{2}
\label{eq:msavi}
\end{equation}

\begin{equation}
MTVI =\frac{1.5 * (1.2 * (N - G) - 2.5 * (R - G))}{\sqrt{(2 * N + 1)^2 - (6 * N - 5 * \sqrt{R}) - 0.5}}
\label{eq:mtvi}
\end{equation}

\begin{equation}
VARI =\frac{G - R}{G + R - B}
\label{eq:vari}
\end{equation}

\begin{equation}
TGI =\frac{(120 * (R - B)) - (190 * (R - G))}{2}
\label{eq:tgi}
\end{equation}

where R, G, B, and N represents the value of the pixel in Red, Green, Blue, and NIR band.

\subsection{Sentinel-1 SAR Imagery}

Since employing the 10-m resolution dual-polarization Ground Range Detected (GRD) scene (VV + VH) aboard Sentinel-1 (S1) has been shown to be useful for several crop mapping related applications \cite{o2020improved, park2018classification, lasko2018mapping, singha2019high}, it was utilized. S1 is an active Synthetic Aperture Radar (SAR) acquiring data in C-band and is not dependent on the time of the day or the weather. Both S1A and S1B were used with both ascending and descending orbit paths. The combined satellite constellation of S1A and S1B provides a revisit time of 6 days, while a single satellite has a revisit time of 12 days. 

The S1 GRD data sets were obtained as an image collection from the GEE. The data set in GEE is pre-processed using the Sentinel-1 SNAP Toolbox \cite{snap_toolbox}. The S1 data set in GEE has gone through a number of pre-processing steps including applications of orbit file, removal of thermal noise, removal of border noise, radiometric calibration, followed by the Geometric Terrain Correction (GTC) using the Shuttle Radar Topography Mission (SRTM) \cite{farr2007shuttle} based Digital Elevation Model (DEM) \cite{markert2020comparing}. Since the SAR image collection in the GEE doesn't perform the Radiometric Terrain Correction (RTC) \cite{small2011flattening}, the RTC corrections in the GEE were performed using the angular based method developed by Vollrath et al \cite{vollrath2020}. Finally, a Lee-sigma speckle filtering \cite{lee2008improved} was applied to further reduce the noise.

\subsection{Other Remote Sensing data sets}
In addition, the Shuttle Radar Topography Mission (SRTM) V3 based digital elevation data were obtained from the GEE. The V3 has undergone a void-filing process using the Advanced Space-borne Thermal Emission and Reflection Radiometer (ASTER) GDEM2 \cite{tachikawa2011characteristics}, Global Multi-resolution Terrain Elevation Data 2010 (GMTED2010) \cite{danielson2011global}, and the National Elevation data set (NED) \cite{gesch2002national}.

\subsection{Deep Learning Algorithms: Neural Network Architecture}
\begin{figure}[hbt]
\centering
\includegraphics[width=\columnwidth]{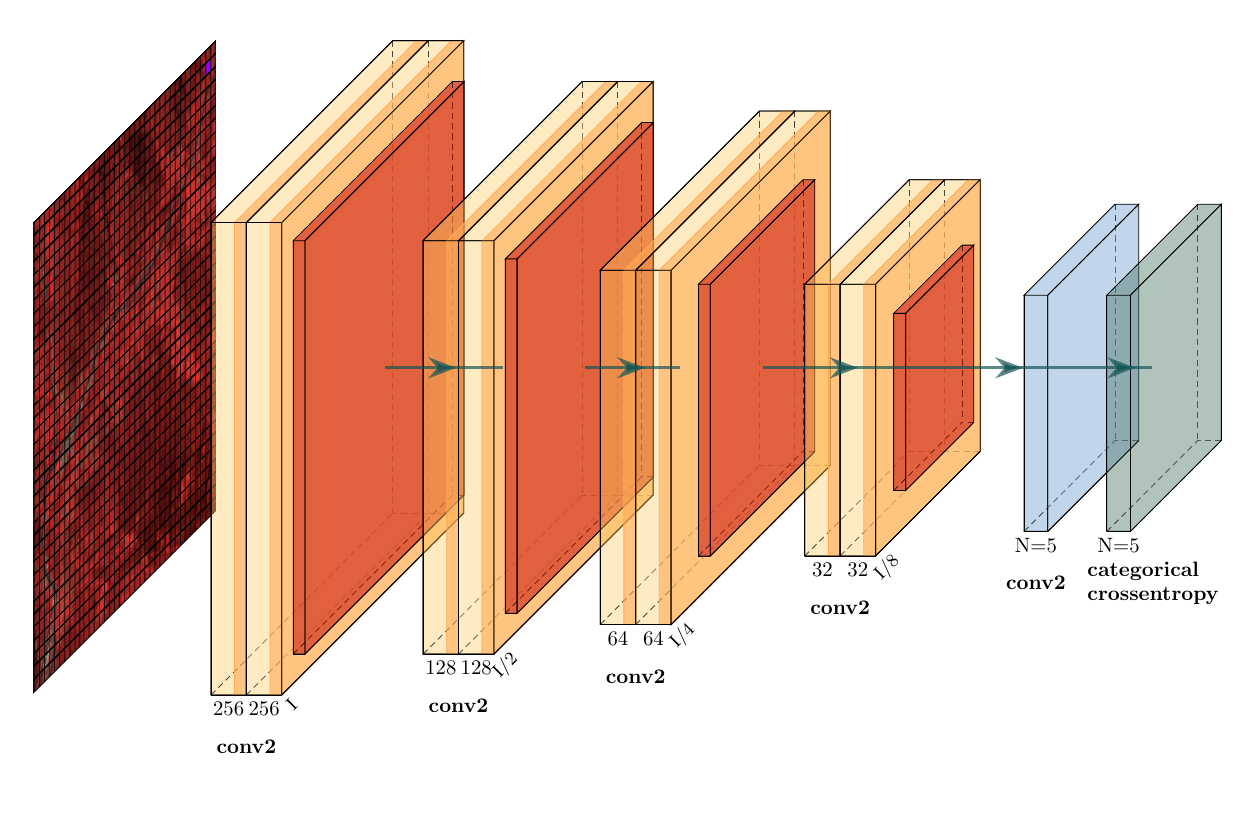} 
\caption[The Deep Neural Network (DNN) model architecture used to map rice extent. The network consists of 1×1 convolution layers (light orange), activation layers (dark orange), max pooling layers (red), and an output layer (light green).] {The Deep Neural Network (DNN) model architecture used to map rice extent in Paro. The network consists of 1×1 convolution layers (light orange), activation layers (dark orange), max pooling layers (red), 2D up-sampling layers (green), and an output layer (light green).}
\label{fig:dnn_arch}
\end{figure}

Two neural network architecture were utilized in this study. The first is the simple Deep Neural Networks (DNN) multilayer perceptron with three multiple hidden layers between input and output. Important to note this DNN approach utilized a pixel-based input data set. For the model architecture a dropout layer was introduced between the dense layers to avoid overfitting. A Rectified Linear Unit (ReLU) for each of these fully connected neural layers was utilized as the activation function. This is also shown in the figure \ref{fig:dnn_arch} \cite{haris_iqbal_2018_2526396}. For the purpose of inference in GEE, a \verb|Conv2D| layer from the \verb|keras| \cite{chollet2015keras} library was used, with 1x1 convolution window resulting in composing this as a Dense layer.

\begin{figure}[hbt]
\centering 
\includegraphics[width=\columnwidth]{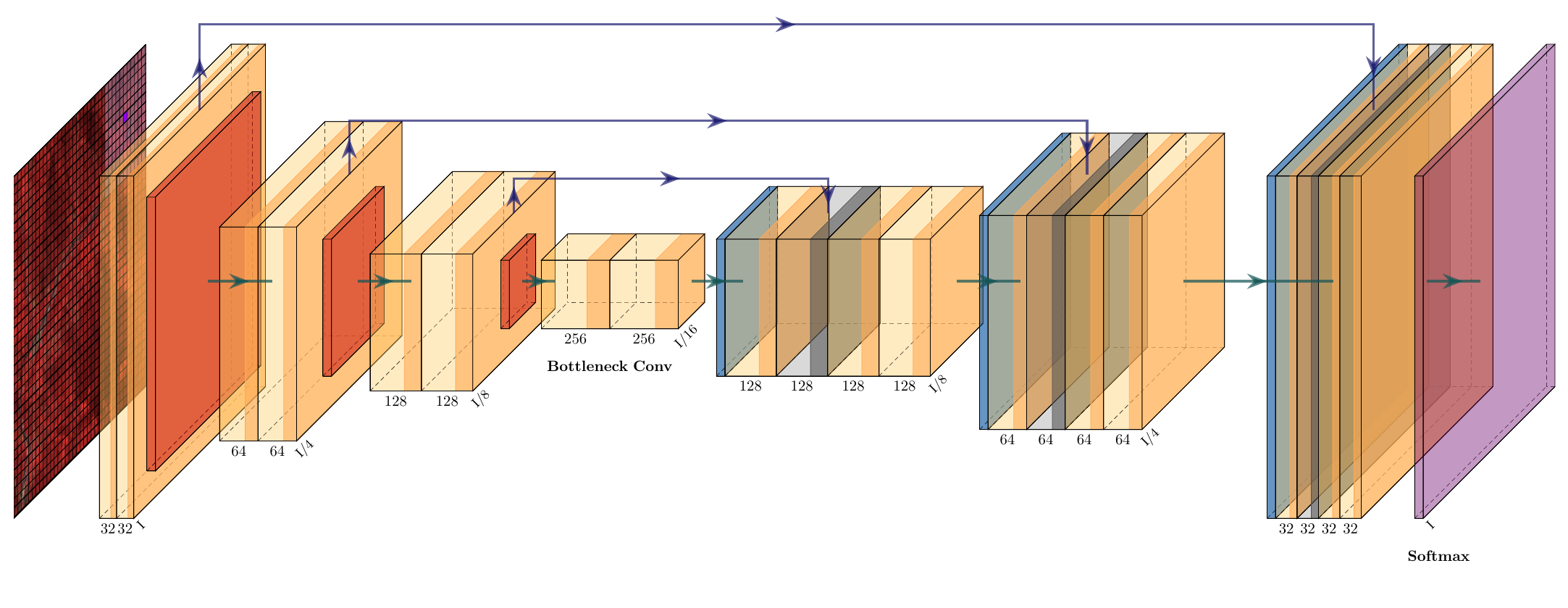} 
\caption[The U-Net model architecture used to map rice extent. The network consists of 3×3 convolution layers (light orange), activation layers (dark orange), max pooling layers (red), 2D up-sampling layers (light blue), and an output layer from the final activation layer (magenta).]{The U-Net model architecture used to map rice extent. The network consists of 3×3 convolution layers (light orange), activation layers (dark orange), max pooling layers (red), 2D up-sampling layers (light blue), and an output layer from the final activation layer (magenta).}
\label{fig:unet_arch}
\end{figure}

Next model that was utilized was the U-Net architecture. U-Net is an encoder decoder architecture model that forms the shape of U \cite{ronneberger2015u}. This U-Net model requires a patch-based input data set, which was set to 256x256. With U-Net models the image space gets reduced while the feature space gets expanded during the encoder stage of the U-Net. And conversely, in the decoder stage image space then gets expanded while the feature space is reduced. The U-Net model consisting of four multiple convolution layer encoding blocks with a distinct max pooling layer at the end of each block. For the convolution layer in the encoding, a depth-wise separable convolutions layer was used. This depth-wise separable convolutions is employed similar to the MobileNetV2 network \cite{sandler2018mobilenetv2}, which helps to significantly decrease the number of operations and memory needed while retaining the same accuracy when using a traditional convolution layer. This is also shown in the figure \ref{fig:unet_arch} \cite{haris_iqbal_2018_2526396}. A softmax activation function was used at the output layer to get a vector of probabilities, representing a probability distribution over the output classes. In addition, the Adam optimizer \cite{kingma2017adam} was used with default settings. Data augmentation was applied that included flipping, rotating, random brightness, and random contrast on the data patches \cite{shorten2019survey}. Data augmentation operations were randomly applied to 80\% of the input data. Furthermore, at specific levels, a residual block through the element-wise addition operations was added to combine the corresponding layer in the encoder and decoder network \cite{ronneberger2015u, he2016deep}.

\subsection{Training Data Sampling}
The training data sampling focused on generating the labeled samples (both points and patches) needed for the model training. Since the dominant land cover in Paro is forest making 47.14\% while agriculture makes only about 5.35\%, there is a high likelihood of a class imbalance problem in this area. As a result, the sampling was focused on the main valley of Paro. This is shown in figure~\ref{fig:study_area_map}. In addition, the weak labels for other classes (besides rice) from the regional model - RLCMS \cite{uddin2021regional} were used in constructing the continuous label to capture the variations between the different strata classes that are especially needed for the patch-based methods like U-Net.

\begin{figure}[hbt]
\centering 
\includegraphics[width=\columnwidth]{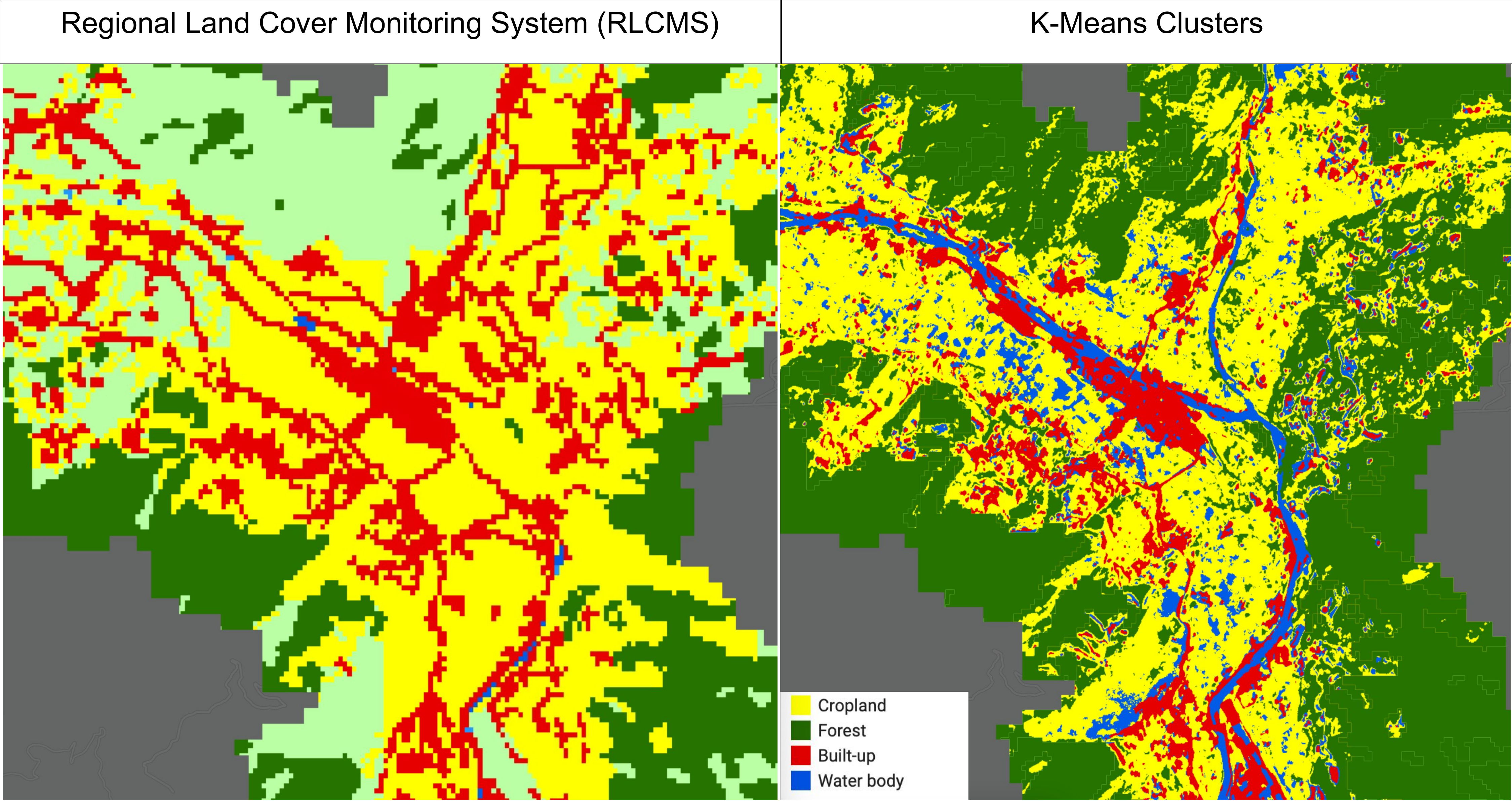} 
\caption[A side-by-side visual comparison between the output of the Regional Land Cover Monitoring System (RLCMS) and K-Means Clusters (after remapping). The clusters from Planet were remapped to resembling class from RLCMS to produce higher spatial resolution land cover map for the sampling purpose.]{A side-by-side visual comparison between the output of the Regional Land Cover Monitoring System (RLCMS) and K-Means Clusters (after remapping). The clusters from Planet were remapped to resembling class from RLCMS to produce higher spatial resolution land cover map for the sampling purpose.}
\label{fig:rlcms_k_means}
\end{figure}

The initial training data sampling steps consists of producing unsupervised clusters of pixels using K-Means clustering \cite{hartigan1979algorithm}. The K-Means clustering is an iterative unsupervised algorithm that divides the data into K number of clusters by minimizing the distance between the cluster center and the data points \cite{lloyd1982least}. A total of seven different clusters were initially produced in GEE. Then the RLCMS was used to compare the closest class to the clusters, which were then manually inspected to generate labels from the RLCMS as the target class, and the clusters were remapped. This resulted in four classes: cropland (including rice), forest, built-up, and others (including water bodies). This is also shown in the figure \ref{fig:rlcms_k_means}.

The RLCMS is a regional model without the specific crop type information; further manual digitized of rice paddies as a separate class was conducted in GEE. This thereby resulted in five classes label: rice, cropland (excluding rice), forest, built-up, and other (including water body). In addition, for the built-up class, an additional more recent building footprints data from the Microsoft was supplemented \cite{MSBuilding2023}. Finally, a total of 12,157 data points were randomly distributed across the sampling area (figure ~\ref{fig:study_area_map}(b)) of which 1030 points consists of cropland, 4332 points were rice class, 2230 points were forest class, 2682 points were built-up class and 1883 points were the other class.

As mentioned prior, DNN models require point-based data while the U-Net models rely on patch-based data inputs. Additionally the U-Net's model performance is largely tied to the spatial variation observed within the image patches \cite{soares2020landslide}. To align with the needed input data structure 12,157, 1x1 patches for the DNN model, and 12,157 256x256 patches, were created with the centers being the random sample point, for the U-Net model. Additionally, the U-Net patches delineated spatial variations across the five classes. Both DNN and U-Net training data sets were exported. The final training data for both DNN and U-Net input data sets were subset into three portions, $\sim$70\% was used for model training, $\sim$20\% for model validation, and $\sim$10\% for final model testing.

\subsection{Performance evaluation}
Categorical accuracy and F1-score (eq (\ref{eq:f1})) were used as the metric for model performance in this study. The categorical accuracy is very similar to the accuracy (see eq. \ref{eq:accuracy}), but instead calculates the percentage of predicted values matching with actual values for one-hot labels. The F1-score is calculated using the precision and recall. The precision (eq (\ref{eq:precision})) is the ratio of correctly predicted positive observations to the total predicted positive observations. The recall (eq (\ref{eq:recall})), also referred to as sensitivity, represents the ratio of correctly predicted positive observations to all the observations in the class. The F1-score (eq (\ref{eq:f1})) \cite{van1979information, chicco2020advantages} is a weighted average of precision and recall. It takes into account both the false positives and false negatives.

\begin{equation}
accuracy =\frac{TN + TP}{TP + FP + TN + FN}
\label{eq:accuracy}
\end{equation}

\begin{equation}
F1 =\frac{2* (recall * precision)}{recall + precision}
\label{eq:f1}
\end{equation}

\begin{equation}
recall =\frac{TP}{TP + FN}
\label{eq:recall}
\end{equation}

\begin{equation}
precision =\frac{TP}{TP + FP}
\label{eq:precision}
\end{equation}

where:

TP is the True Positives, which means that the actual class and the predicted class are both positive.

TN is the True Negatives, which means that the actual and predicted class are both negative.

FP is the False Positives, which means that the actual class is negative whereas the predicted class is positive.

FN is the False Negative, which means that the actual class is positive but the predicted class is negative. \\

Beyond the model validation, an additional separate independent validation was conducted. A random sampling approach was used for comparisons of all the model outputs. Initially, 1667 random stratified points were generated across Paro and added to an independent Collect Earth Online (CEO) \cite{saah2019collect} survey project. These 1667 30-m plots  were manually interpreted by trained samplers and labeled (classified) using the 2021 NICFI monthly mosaic imagery in CEO. Each plot was labeled rice, non-rice, or mixed class. Out of the 1667 random stratified sampled plots, 359 points were within the sampled geometry (See figure~\ref{fig:study_area_map} (c)) and therefore were excluded from the final model independent validation.

\subsection{Modeling Approaches}

\begin{figure}[hbt]
\centering 
\includegraphics[width=\columnwidth]{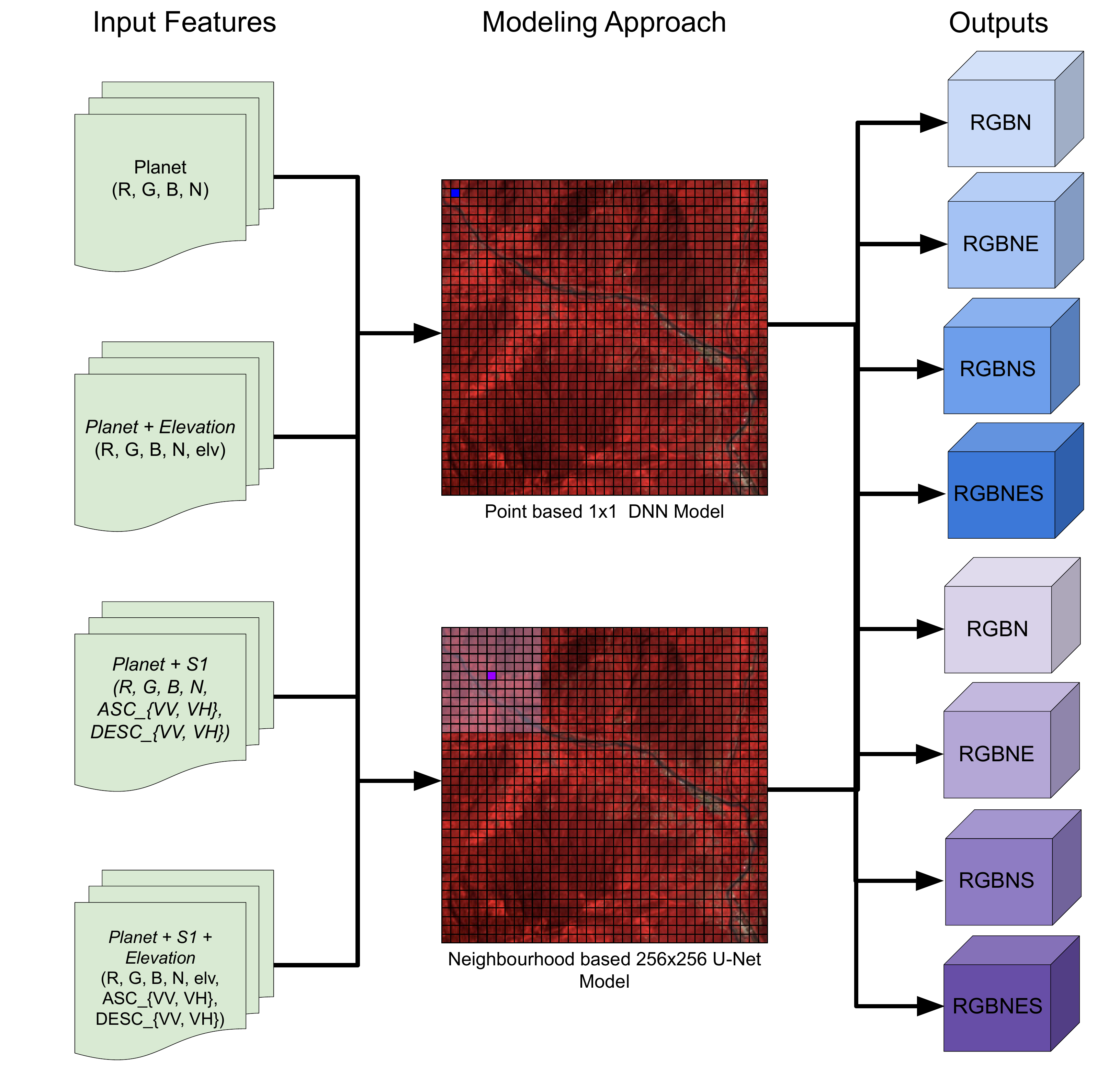} 
\caption[Overview of the study design for the distinct architecture approaches (DNN and U-Net) and training feature combinations (RGBN, RGBNE, RGBNS, RGBNES)]{Overview of the study design for the distinct architecture approaches (DNN and U-Net) and training feature combinations (RGBN, RGBNE, RGBNS, RGBNES)}
\label{fig:model_comparison_approaches}
\end{figure}

Both a pixel-based DNN and patch-based U-Net models were trained using four variations of the input training feature data sets being 1) Planet images only, 2) combining Planet images with elevation data, 3) combining Planet images with S1 images, and 4) combining Planet images with both elevation data and S1 images, which are respectively referred to as RGBN, RGBNE, RGBNS, RGBNES respectively hereafter. Between the two DL approaches, a total of eight model combinations were derived. This is also shown in figure ~\ref{fig:model_comparison_approaches}.

\begin{figure}[hbt]
\centering 
\includegraphics[width=\columnwidth]{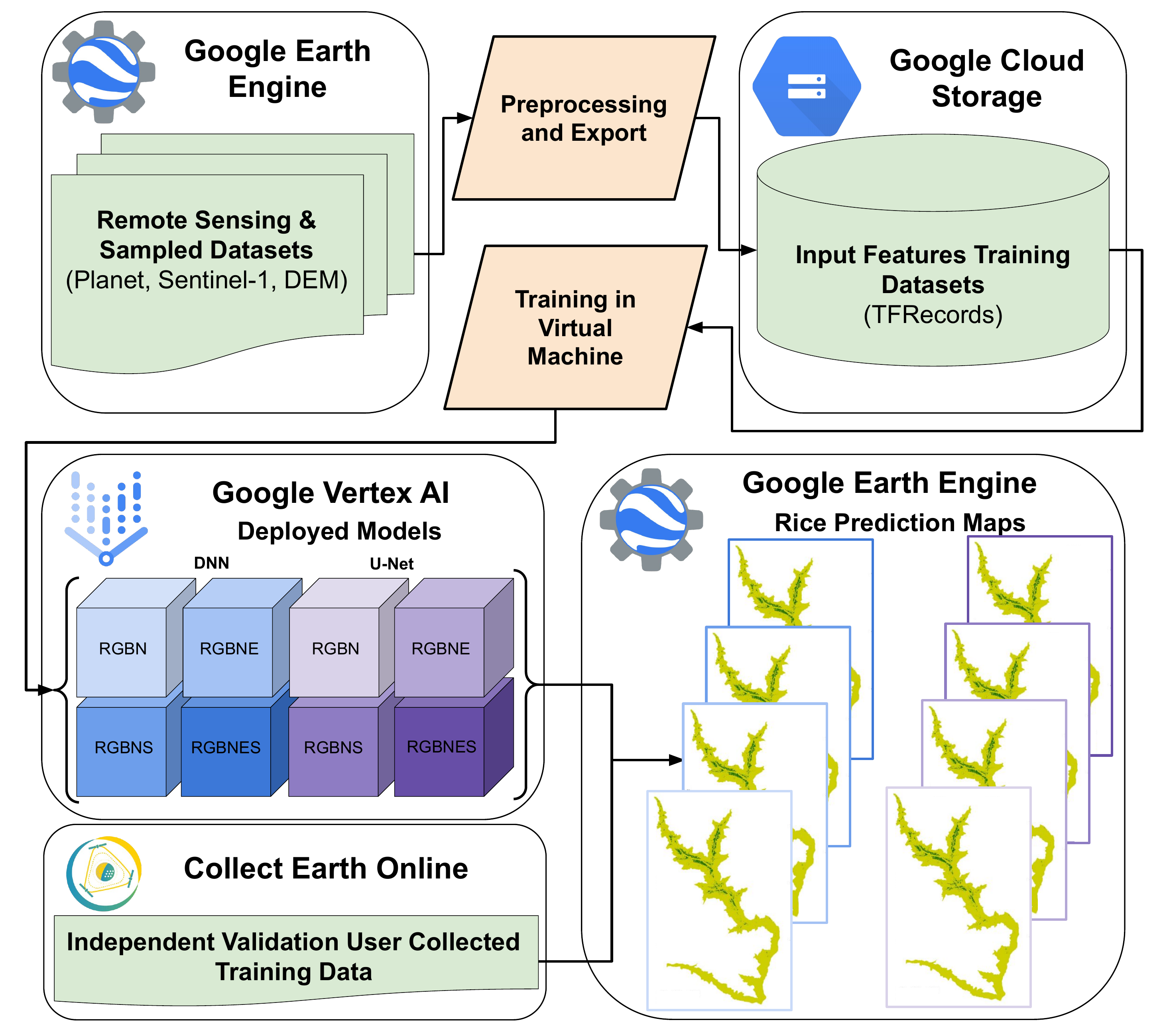}
\caption[Processing workflow for the generation of the rice maps. The platform where each individual processes took place are shown with the logo on the top.]{Processing workflow for the generation of the rice maps. The platform where each individual processes took place are shown with the logo on the top.}
\label{fig:workflow}
\end{figure}

The overall workflow included pre-processing and generating sample points and training patches using the GEE platform. The training data were then exported as a TFRecords. The TensorFlow library \cite{tensorflow2015-whitepaper} was used for the DL training purpose in a local machine. Once an individual model was trained, it was deployed to the Vertex AI platform as an endpoint. The Vertex AI platform is the overall Machine Learning (ML) platform developed by Google to train and deploy the ML and DL models. Then the inference on the deployed endpoint was performed using the GEE. This comprehensive workflow is also presented in the figure ~\ref{fig:workflow}.

\section{Preliminary Testing}
Since DNN and UNET are able to learn and derive features with their dense network architecture, preliminary testing was conducted to see the differences in the performance when using only the core RGBN bands of the Planet data and by adding RGBN with other derived features. These derived features were NDVI, EVI, NDWI, SAVI, MSAVI, MTVI, VARI, and TGI. The formulas to calculate each derived feature are provided in eqs \ref{eq:ndvi}, \ref{eq:evi}, \ref{eq:ndwi}, \ref{eq:savi}, \ref{eq:msavi}, \ref{eq:mtvi}, \ref{eq:vari}, and \ref{eq:tgi} respectively. For this preliminary analysis, the Planet imagery selected were used from before and during the growing season (June-September).

\begin{figure}[hbt]
\centering 
\includegraphics[width=\columnwidth]{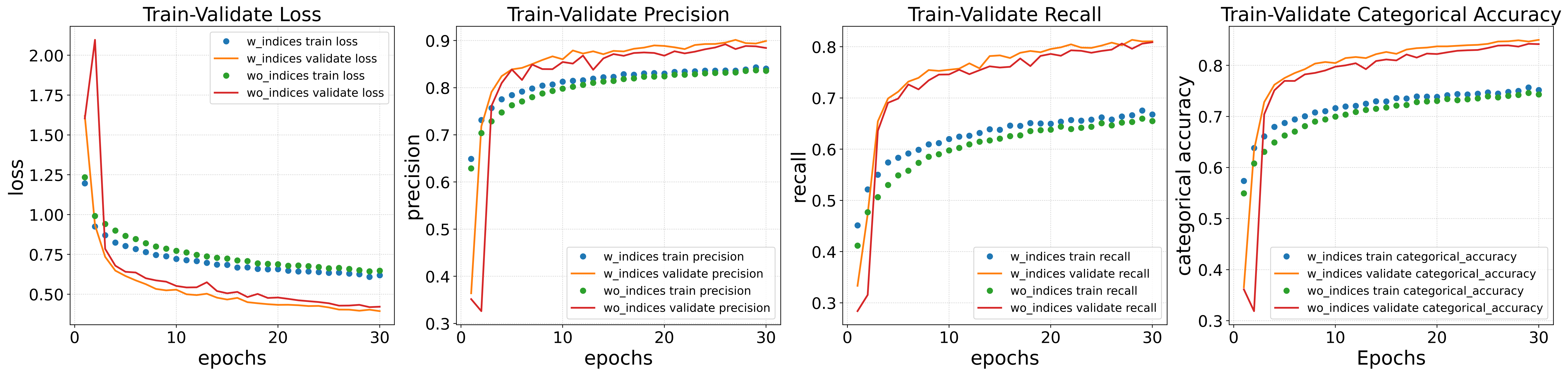}
\caption[Training and validation loss of using a U-Net Model with using indices as additional features in addition to RGBN and without using the indices.]{Training and validation loss of using a U-Net Model with using indices as additional features in addition to RGBN and without using the indices.}
\label{fig:model_w_wo_indices} 
\end{figure}

The preliminary results of training the U-Net model for 30 epochs is shown in figure ~\ref{fig:model_w_wo_indices} and in tabular format in Table~\ref{tab:model_w_wo_indices}. As shown in the table, across all the evaluation metrics (validation precision, recall, and categorical accuracy), both preliminary testing sets perform similarly. When evaluating the metrics for the training test data, which was previously unobserved by the model, these metrics display a F1 score of 85.21\% and 84.44\% (and high precision, recall, and accuracy) for when including indices and excluding indices respectively. From this preliminary testing models, all models produced utilized only the core RGBN bands as not much increased performance was observed with additional feature engineering with derived indices.

\begin{table}[H]
\caption{The following table provides the comparison of using a U-Net Model with using indices as additional features in addition to RGBN and without using the indices.}
\label{tab:model_w_wo_indices}
\begin{tabular}{llccll}
\hline
 &
  \multicolumn{1}{c}{\textbf{Loss}} &
  \textbf{Categorical Accuracy} &
  \textbf{Precision} &
  \multicolumn{1}{c}{\textbf{Recall}} &
  \multicolumn{1}{c}{\textbf{F1}} \\ \hline
Model with Indices &
  0.3948 &
  0.8495 &
  0.8987 &
  0.8101 &
  0.8521 \\ \hline
Model without Indices &
  0.4224 &
  0.8412 &
  0.8841 &
  0.8081 &
  0.8444 \\ \hline
\end{tabular}
\end{table}

\section{Results and Discussion}

\subsection{U-Net Result}
\begin{figure}[hbt]
\centering 
\includegraphics[width=\columnwidth]{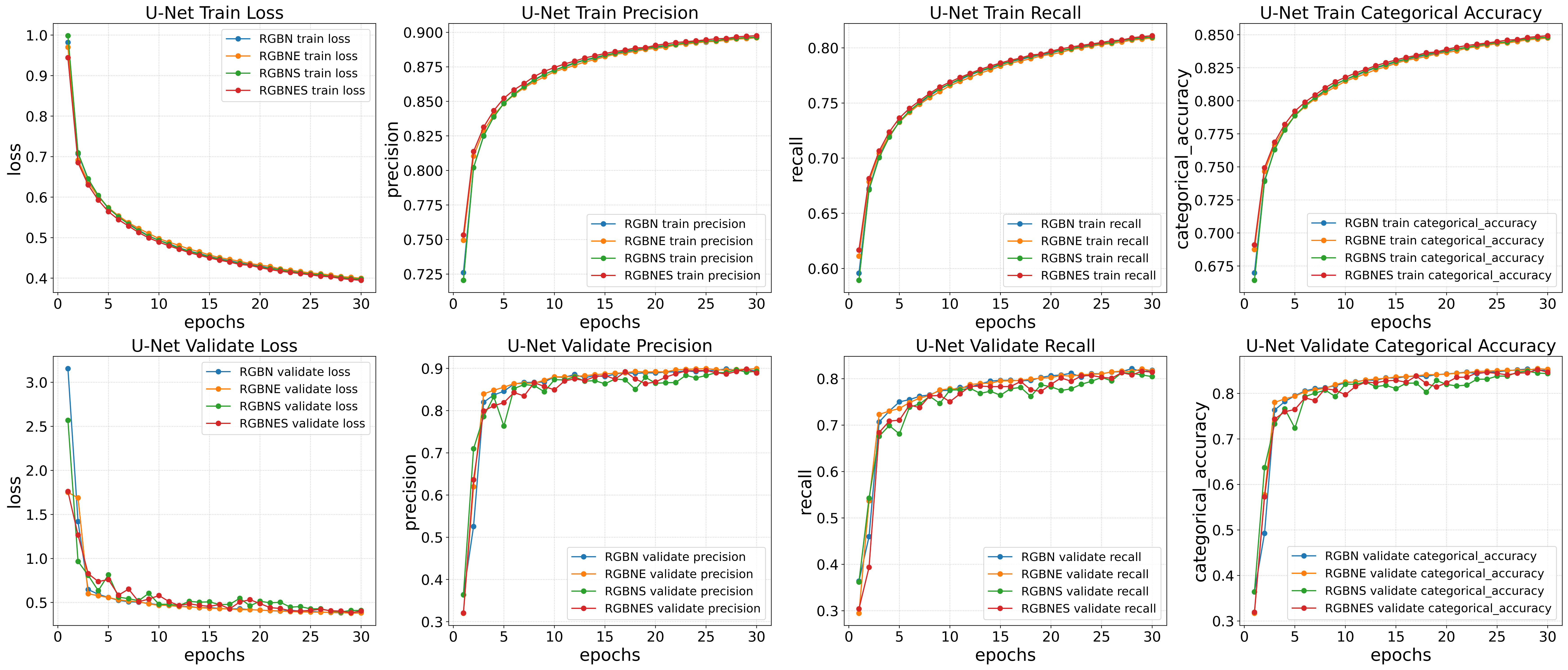} 
\caption[Training and validation plots for the process of training a U-Net model with using 1) Planet images only, 2) combining Planet images with elevation data, 3) combining Planet images with Sentinel-1 data, 4) combining Planet images with elevation and Sentinel-1 data.]{Training and validation plots of training a U-Net Model with using using Planet images only, with combining Planet images with elevation data, with combining Planet images with Sentinel-1 data, and with combining Planet images with elevation and Sentinel-1 data.}
\label{fig:metrics_plot_unet_model_comparison}
\end{figure}

Figure~\ref{fig:metrics_plot_unet_model_comparison} shows the results of applying a trained U-Net model with 1) Planet images, 2) combining Planet images with elevation data, 3) combining Planet images with the Sentinel-1 data, and 4) combining Planet images with Sentinel-1 and elevation data, which are referred to as U-Net:RGBN, U-Net:RGBNE, U-Net:RGBNS, and U-Net:RGBNES respectively. The final evaluation score on the test data set for the different combination of the models are displayed in table~\ref{table:u-net_model_results}. From the analysis, combining Planet with elevation data (U-Net:RGBNE) displayed the best relative performance (0.8532) followed by U-NET:RGBN (0.8512), U-Net:RGBNES (0.8467), and U-Net:RGBNS (0.8440) when ranking via the categorical accuracy metric across the four trial sets. However extremely close values across all of the metrics including precision, recall, and F1 scores were observed expressing a strong performance by the U-Net modeling approach.

\begin{table}[H]
\caption[The following table provides the result of applying the U-Net model with test data set with various combinations of using Planet images only and with elevation and Sentinel-1]{The following table provides the result of applying the DNN model with test data set with various combinations of using Planet images only and with elevation and Sentinel-1}

\label{table:u-net_model_results}
\begin{tabular}{cccccc}
\hline
\textbf{} & \textbf{Loss} & \textbf{\begin{tabular}[c]{@{}c@{}}Categorical\\ Accuracy\end{tabular}} & \textbf{Precision} & \textbf{Recall} & \textbf{F1} \\ \hline
U-Net:RGBN             & 0.3873 & 0.8512 & 0.8943 & 0.8182 & 0.8546 \\ \hline
U-Net:RGBNE     & 0.3814 & 0.8532 & 0.8997 & 0.8169 & 0.8563 \\ \hline
U-Net:RGBNS           & 0.4088 & 0.8440 & 0.8934 & 0.8046 & 0.8467 \\ \hline
U-Net:RGBNES & 0.4037 & 0.8467 & 0.8892 & 0.8141 & 0.8500 \\ \hline
\end{tabular}
\end{table}

\subsection{DNN Result}
\begin{figure}[hbt]
\centering 
\includegraphics[width=\columnwidth]{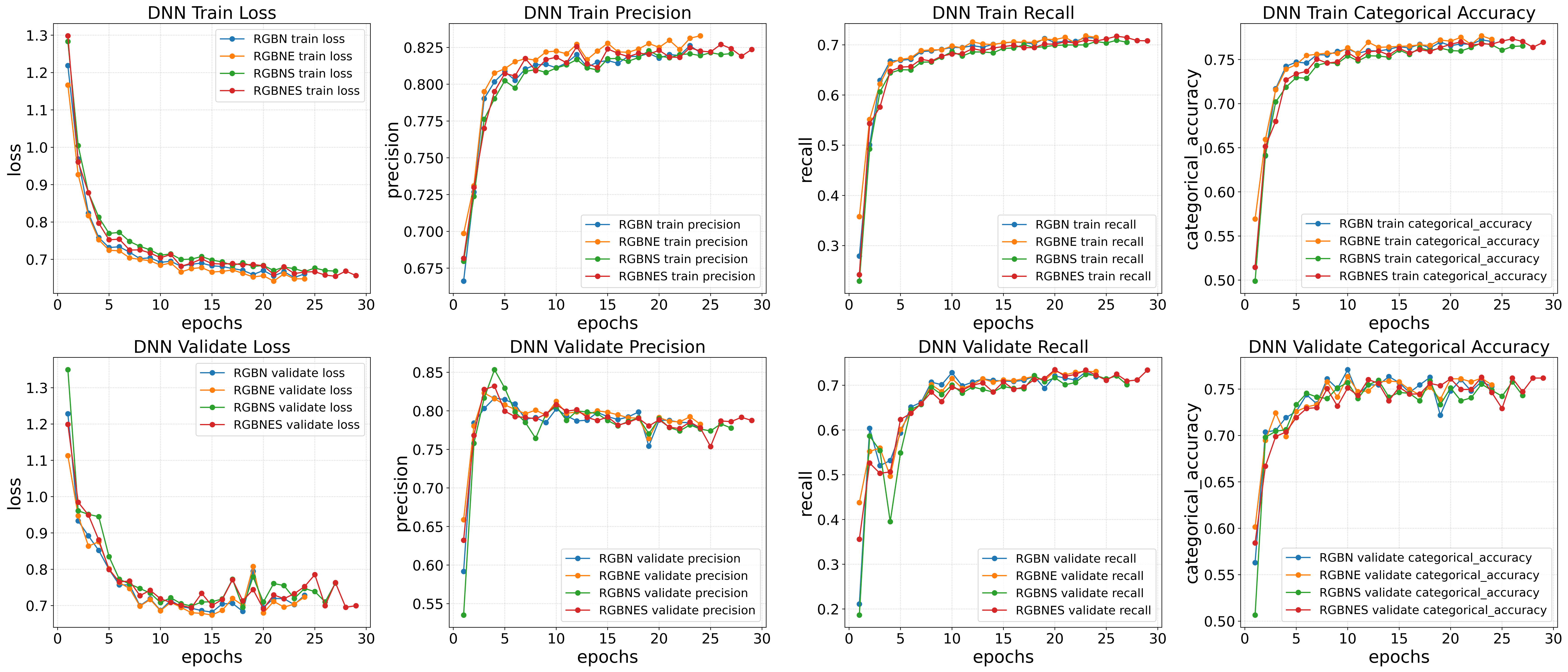} 
\caption[The following table provides the result of applying the DNN model with test data set with various combinations of using Planet images only and with elevation and Sentinel-1]{The following table provides the result of applying the DNN model with test data set with various combinations of using Planet images only and with elevation and Sentinel-1}
\label{fig:metrics_plot_dnn_model_comparison}
\end{figure}

Similarly, Figure~\ref{fig:metrics_plot_dnn_model_comparison} shows the results of applying a trained DNN model with the four distinct input training data sets DNN:RGBN, DNN:RGBNE, DNN:RGBNS, DNN:RGBNES. The final evaluation score on the test data sets for the different combination of the models are also presented in table~\ref{table:dnn_model_results}. Similar to the U-Net approach, the model with Planet and Elevation data (DNN:RGBNE) displayed the highest categorical accuracy of (0.7678) followed closely by model DNN:RGBNES (0.7654), then model RGBN (0.7612), and finally model DNN:RGBNS (0.7564). Similarly, very close values can be seen across the F1 scores indicating similar performance.

\begin{table}[H]
\caption{The following table provides the result of applying the DNN model with test data set with various combinations of using Planet images only and with elevation and Sentinel-1.}
\label{table:dnn_model_results}
\begin{tabular}{cccccc}
\hline
\textbf{} & \textbf{Loss} & \textbf{\begin{tabular}[c]{@{}c@{}}Categorical\\ Accuracy\end{tabular}} & \textbf{Precision} & \textbf{Recall} & \textbf{F1} \\ \hline
DNN:RGBN             & 0.6430 & 0.7612 & 0.8126 & 0.7186 & 0.7627 \\ \hline
DNN:RGBNE     & 0.6434 & 0.7678 & 0.8202 & 0.7260 & 0.7702 \\ \hline
DNN:RGBNS           & 0.6882 & 0.7564 & 0.8045 & 0.7260 & 0.7632 \\ \hline
DNN:RGBNES & 0.6927 & 0.7654 & 0.8041 & 0.7408 & 0.7712 \\ \hline
\end{tabular}
\end{table}

\subsection{Independent Validation}
A majority probability class approach was used to convert the probability class vectors into a rice and non-rice binary classification map. The accuracy, precision, recall, and F1-score of the independent validation points are shown in Table \ref{tab:dnn_independent_validation} and \ref{tab:unet_independent_validation}. It should be noted that the independent validation metrics are distinct from the model validation metric reported. The variation observed further outlines the need to practitioners to employ completely separate independent validation efforts. For this independent validation, all the models display a very high accuracy, while the range across precision and recall varies, especially within the DNN models. For example, the model DNN:RGBNS has a very high recall (81.93\%), but it also has the lowest precision (37.57\%) which means that while this model is good at detecting rice, it also over predicts resulting in a high number of false positives. Similarly, the model DNN:RGBN has high precision (72.06\%), but relatively lower recall (59.04\%), signifying this model is conservative in predicting the rice class and predicts positives only on instances that have a high confidence and thereby missing actual positives resulting in more false negatives. While the U-Net:RGBNES, on average has a good balance between the precision and recall while having high accuracy, suggesting it may be the most effective model across all tested.

\begin{table}[H]
\caption{The following table provides the result of the independent validation of using the binary rice and non-rice layer with the DNN model.}
\label{tab:dnn_independent_validation}
\begin{tabular}{ccccc}
\hline
\textbf{Models} & \textbf{\begin{tabular}[c]{@{}c@{}}Accuracy\end{tabular}} & \textbf{Precision} & \textbf{Recall} & \textbf{F1} \\ \hline
DNN:RGBN   & 0.9593 & 0.7206 & 0.5904 & 0.6490 \\ \hline
DNN:RGBNE  & 0.9478 & 0.5789 & 0.6627 & 0.6180 \\ \hline
DNN:RGBNS  & 0.9018 & 0.3757 & 0.8193 & 0.5152 \\ \hline
DNN:RGBNES & 0.9386 & 0.5140 & 0.6627 & 0.5789 \\ \hline
\end{tabular}
\end{table}

\begin{table}[H]
\caption{The following table provides the result of the independent validation of using the binary rice and non-rice layer with the U-Net model.}
\label{tab:unet_independent_validation}
\begin{tabular}{ccccc}
\hline
\textbf{Models} & \textbf{\begin{tabular}[c]{@{}c@{}}Accuracy\end{tabular}} & \textbf{Precision} & \textbf{Recall} & \textbf{F1} \\ \hline
U-Net:RGBN   & 0.9517 & 0.6389 & 0.5542 & 0.5935 \\ \hline
U-Net:RGBNE  & 0.9463 & 0.5657 & 0.6747 & 0.6154 \\ \hline
U-Net:RGBNS  & 0.9463 & 0.5747 & 0.6024 & 0.5882 \\ \hline
U-Net:RGBNES & 0.9586 & 0.6933 & 0.6265 & 0.6582 \\ \hline
\end{tabular}
\end{table}

\begin{figure}[hbt]
\centering 
\includegraphics[width=\columnwidth]{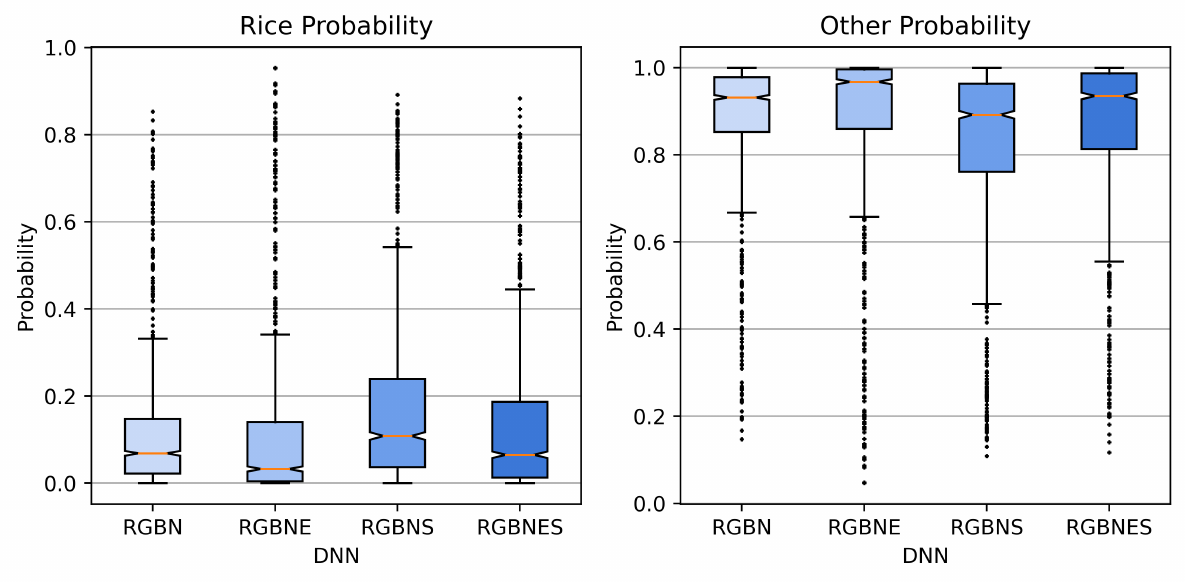} 
\caption[Probability distribution of the independent validation (N = 1308 points) for the DNN models.]{Probability distribution of the independent validation (N = 1308 points) for the DNN models.}
\label{fig:DNN_indep-val-prob}
\end{figure}

\begin{figure}[hbt]
\centering 
\includegraphics[width=\columnwidth]{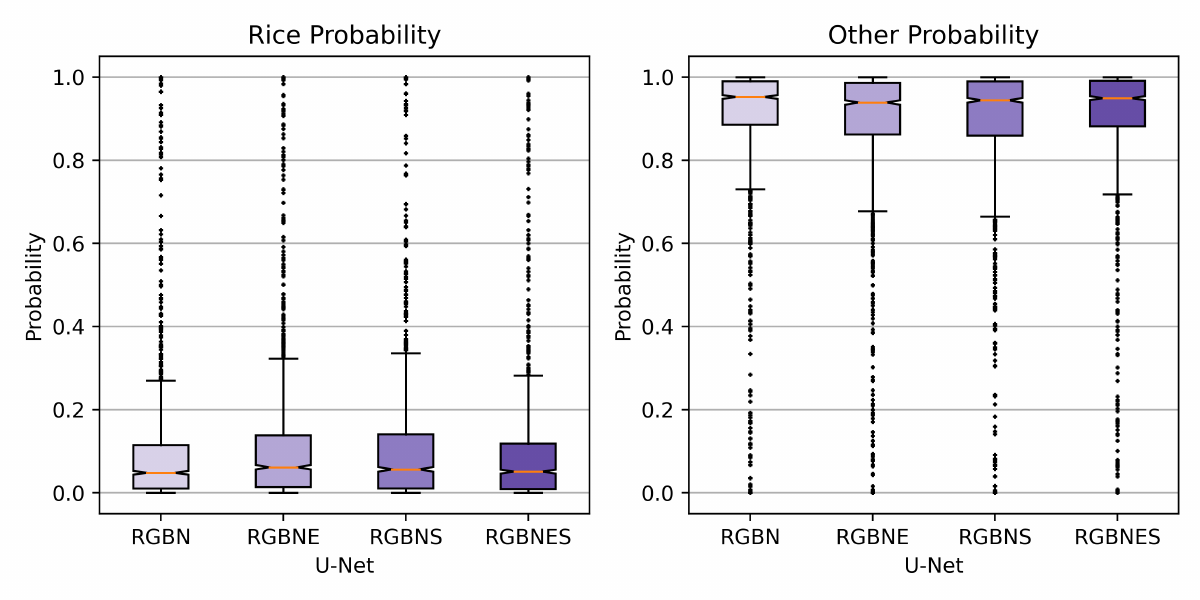} 
\caption[Probability distribution of the independent validation (N = 1308 points) for U-Net models.]{Probability distribution of the independent validation (N = 1308 points) for U-Net models.}
\label{fig:Unet_indep-val-prob}
\end{figure}

Similarly, figures~\ref{fig:DNN_indep-val-prob} and~\ref{fig:Unet_indep-val-prob} both show boxplot of the probability distribution of the independent validation (N=1308 points) for the DNN and U-Net models. When comparing the distributions of the DNN and U-Net model, the DNN model sets displays a larger variation, while U-Net display relatively similar interquartile range and median value for both rice and non-rice proababilities. Additionally, for both DNN and U-Net model sets it can also be seen that nearly all of the rice points are plotted as an outlier, except for DNN:RGBNS which as discussed above overpredicts the rice class. These outlier across the model sets underscore the pervasive class imbalance within the independent validation sample points, whereby majority of the sample points are non-rice (N = 1,225) while only few are the rice class (N = 83), as noted above rice is a relatively spare land cover class within Bhutan and the independent validation data set's distributions parallels this phenomenon.

\subsection{Discussion}

\begin{figure}[!hbt]
\centering 
\includegraphics[width=\columnwidth]{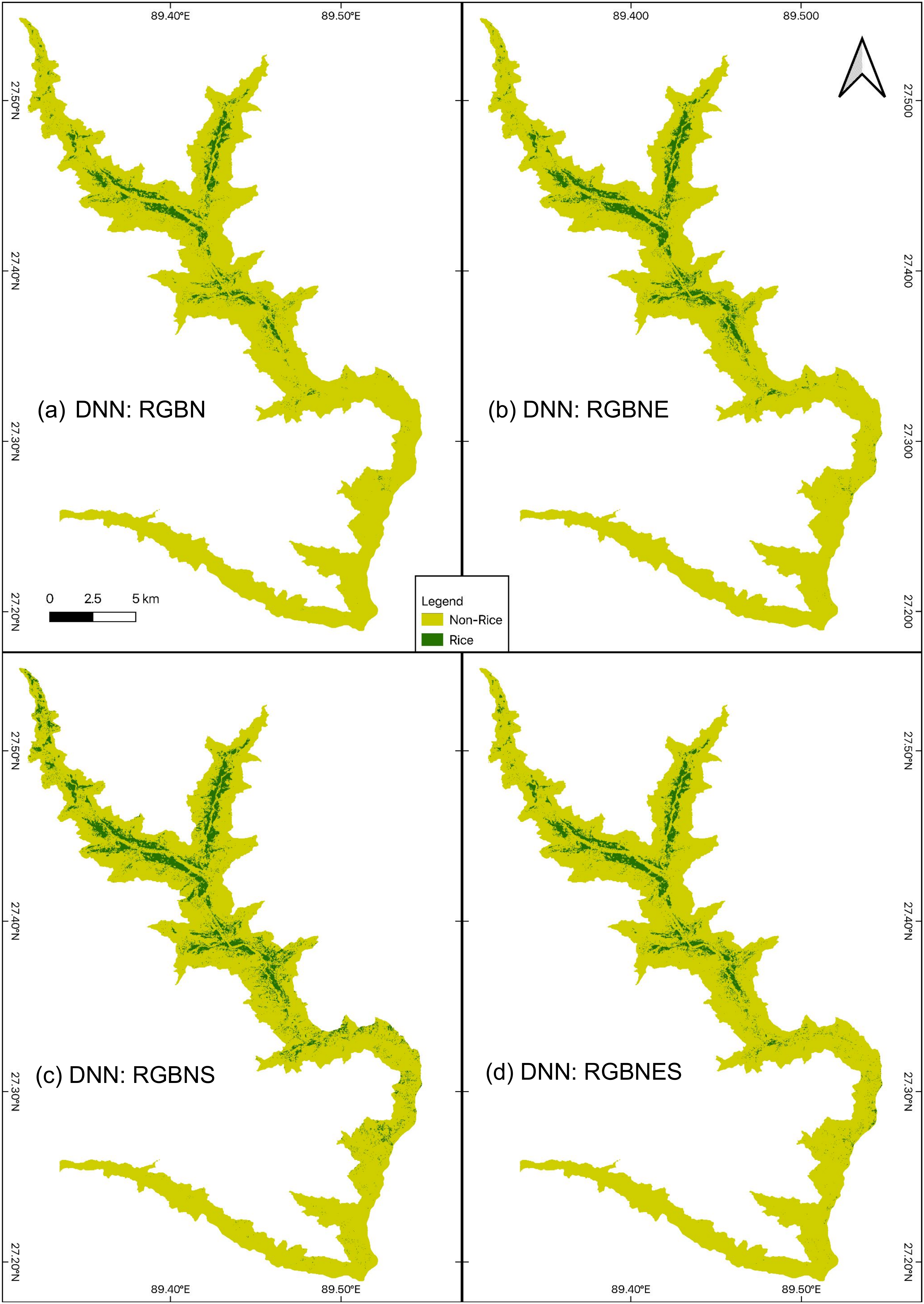} 
\caption[Rice maps for the Paro Dzongkhag for 2021 from DNN model. The output from four different models: RGBN, RGBNE, RGBNS, and RGBNES are shown in sub figure (a), (b), (c) and (d) respectively.]{Rice maps for the Paro Dzongkhag for 2021 from DNN model. The output from four different models: RGBN, RGBNE, RGBNS, and RGBNES are shown in sub figure (a), (b), (c) and (d) respectively.}
\label{fig:dnn_rice_maps}
\end{figure}

\begin{figure}[!hbt]
\centering 
\includegraphics[width=\columnwidth]{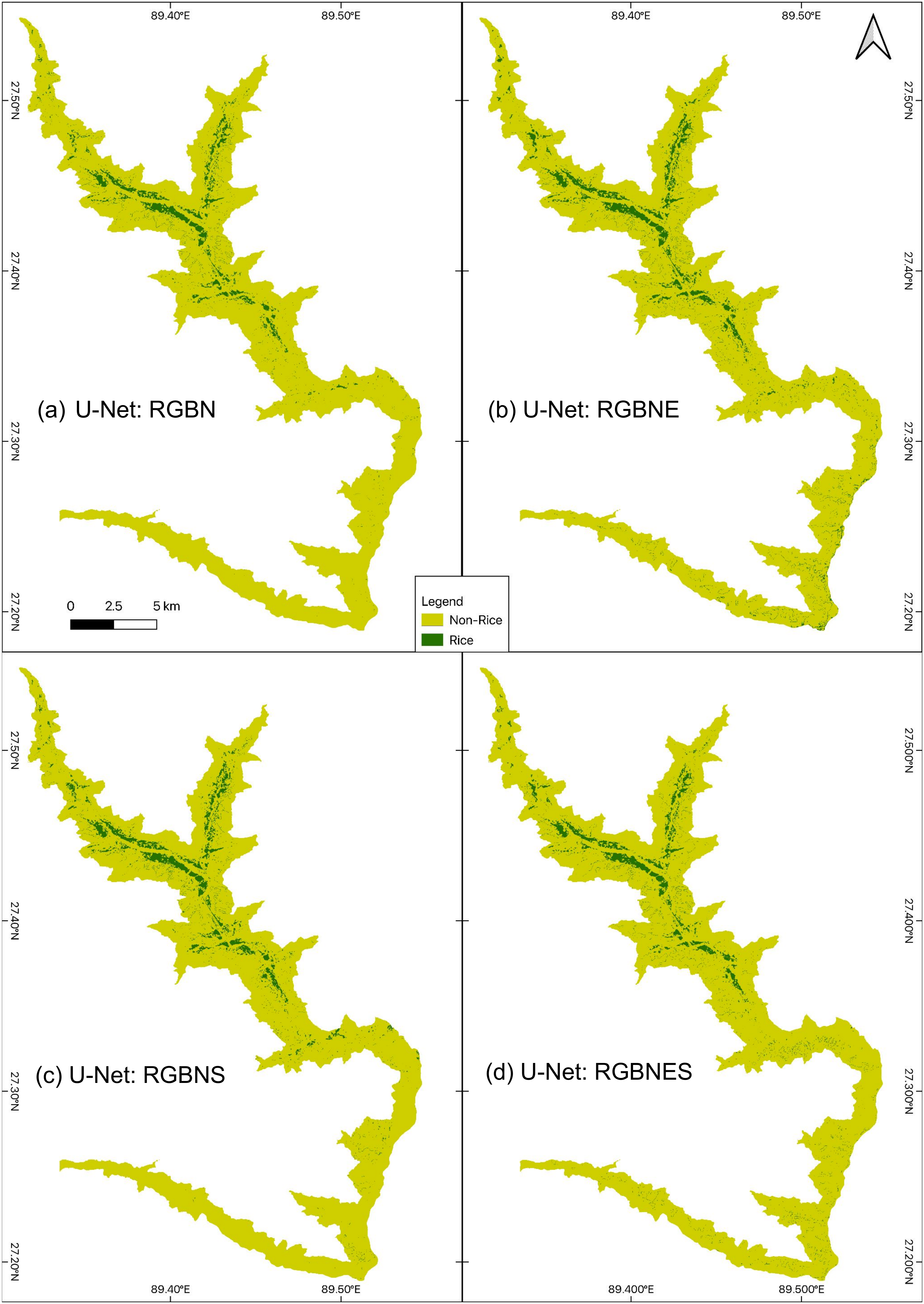} 
\caption[Rice maps for the Paro Dzongkhag for 2021 from U-Net model. The output from four different models: RGBN, RGBNE, RGBNS, and RGBNES are shown in sub figure (a), (b), (c) and (d) respectively.]{Rice maps for the Paro Dzongkhag for 2021 from U-Net model. The output from four different models: RGBN, RGBNE, RGBNS, and RGBNES are shown in sub figure (a), (b), (c) and (d) respectively.}
\label{fig:unet_rice_maps}
\end{figure}

\begin{figure}[hbt]
\centering 
\includegraphics[width=\columnwidth]{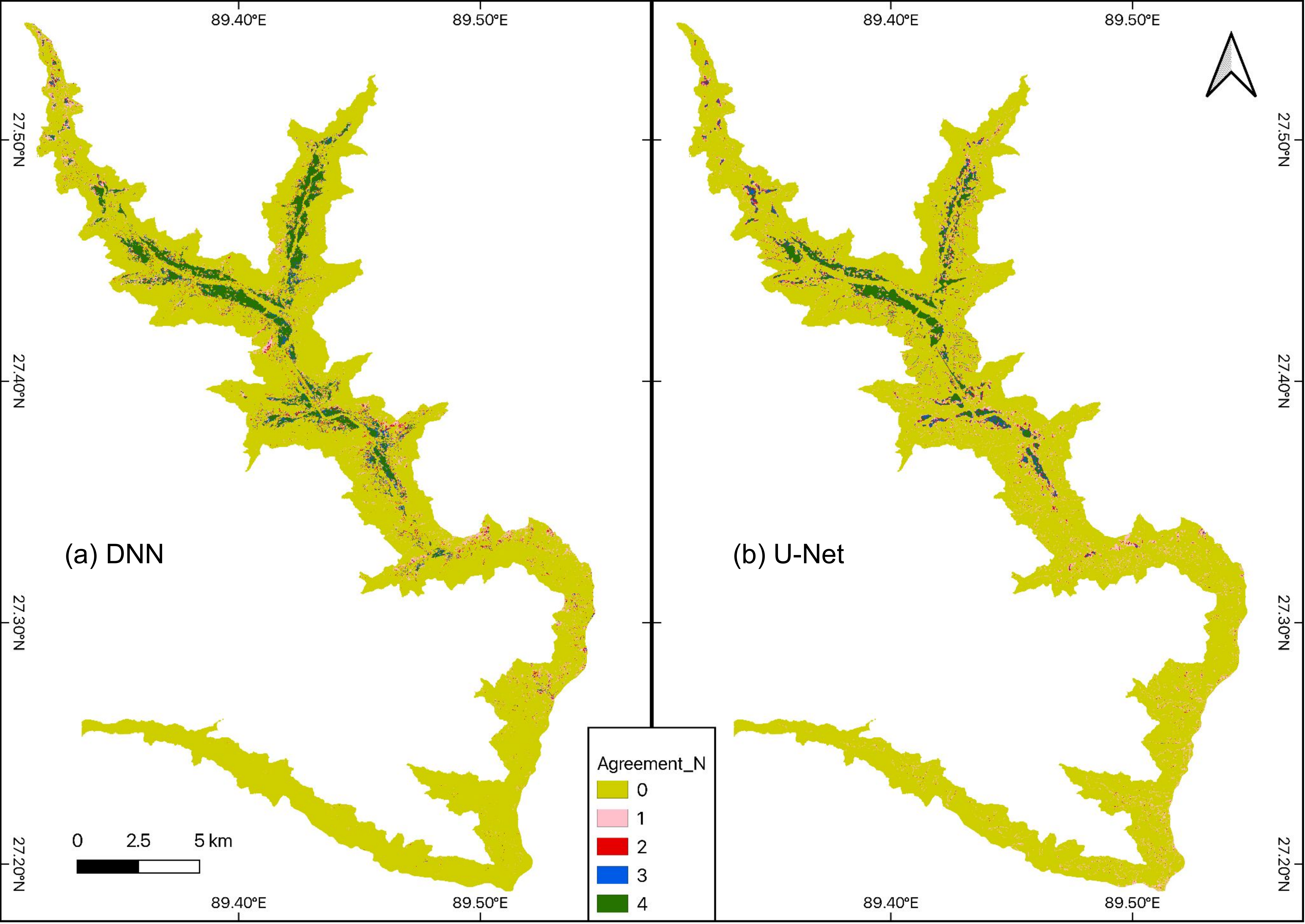} 
\caption[Agreement map for the output for (a) DNN and (b) U-Net. A binary image of rice and non-rice maps were produced, and the sum of all models was produced, where a value of 4 represents that all models predicted rice and 0 represents that 0 models predicted rice.]{Agreement map for the output for (a) DNN and (b) U-Net. A binary image of rice and non-rice maps were produced, and the sum of all models was produced, where a value of 4 represents that all models predicted rice and 0 represents that 0 models predicted rice.}
\label{fig:agreement_maps}
\end{figure}

The results for the produced rice maps utilizing the DNN and U-Net architectures for 2021 in the Paro Dzongkhag are displayed in figure~\ref{fig:dnn_rice_maps} and figure~\ref{fig:unet_rice_maps}. The output from four different models: RGBN, RGBNE, RGBNS, and RGBNES are shown in sub figure (a), (b), (c) and (d) respectively. In addition, a binary image of rice and non-rice maps were produced, and the sum of all models for both DNN and U-Net were produced which is shown in figure~\ref{fig:agreement_maps}. In this agreement analysis, a value of 4 represents all models agree as predicted rice and 0 represents that no models predicted rice. In figure ~\ref{fig:agreement_maps}, both DNN and U-Net depicts model agreement in the main valley floor of Paro where Rice paddies are contiguous and grown densely in relatively large fields. However, the agreement continues to decrease when proceeding southward. As an additional context, the geography and topography and thereby cultivation practices employed in Paro range significantly from North to South and this is observed in the model results. The Southern reach of Paro was not included within the initial training data collect region of interest both due to lower quantity of cultivated rice as well as smaller and less contiguous fields are present in this region.

Additionally, the predicted rice area was calculated and compared with the National Statistics Bureau (NBS) reported annual Agriculture Survey Report of 2021 for Paro. The NBS reported the total sown area of 2,066.9 acres for Paro. The calculated area from all eight different model output are listed in table~\ref{tab:area_calculation}. All the DL based model over predicts the rice paddy area when compared with the number reported by the NBS, however this range of over prediction differs greatly among the models. The lowest difference can be seen with the U-Net:RGBN, with only 9.27\% overestimation, while the DNN:RGBNS over predicts by more than 2.5 times. The overprediction from DNN:RGBNS can also be inferred through independent validation table ~\ref{tab:dnn_independent_validation}, which predicts many false positives as evident by its high recall and low precision. However, it is also important to note that the current methods for the area estimation for the report is based on survey and questionnaire method (total 1018 household surveyed in Paro in 2021) and doesn't use Remote Sensing approaches, as a result some uncertainty may be expected \cite{NSB_Ag_2021}.

\begin{table}[H]
\caption{The following table provides the area calculated from the produced map in acres. DNN over-predicts across every model compared to the U-Net.}
\label{tab:area_calculation}
\begin{tabular}{ccccc}
\hline
\multirow{2}{*}{\textbf{Models}} & \textbf{RGBN} & \textbf{RGBNE} & \textbf{RGBNS} & \textbf{RGBNES} \\ \cline{2-5} 
      & \multicolumn{4}{c}{\textbf{Area in Acres}} \\ \hline
DNN   & 2584.72   & 3471.83  & 5181.80  & 3407.58  \\ \hline
U-Net & 2258.54   & 3197.54  & 2807.12  & 2542.71  \\ \hline
\end{tabular}
\end{table}

\begin{figure}[hbt]
\centering 
\includegraphics[width=\columnwidth]{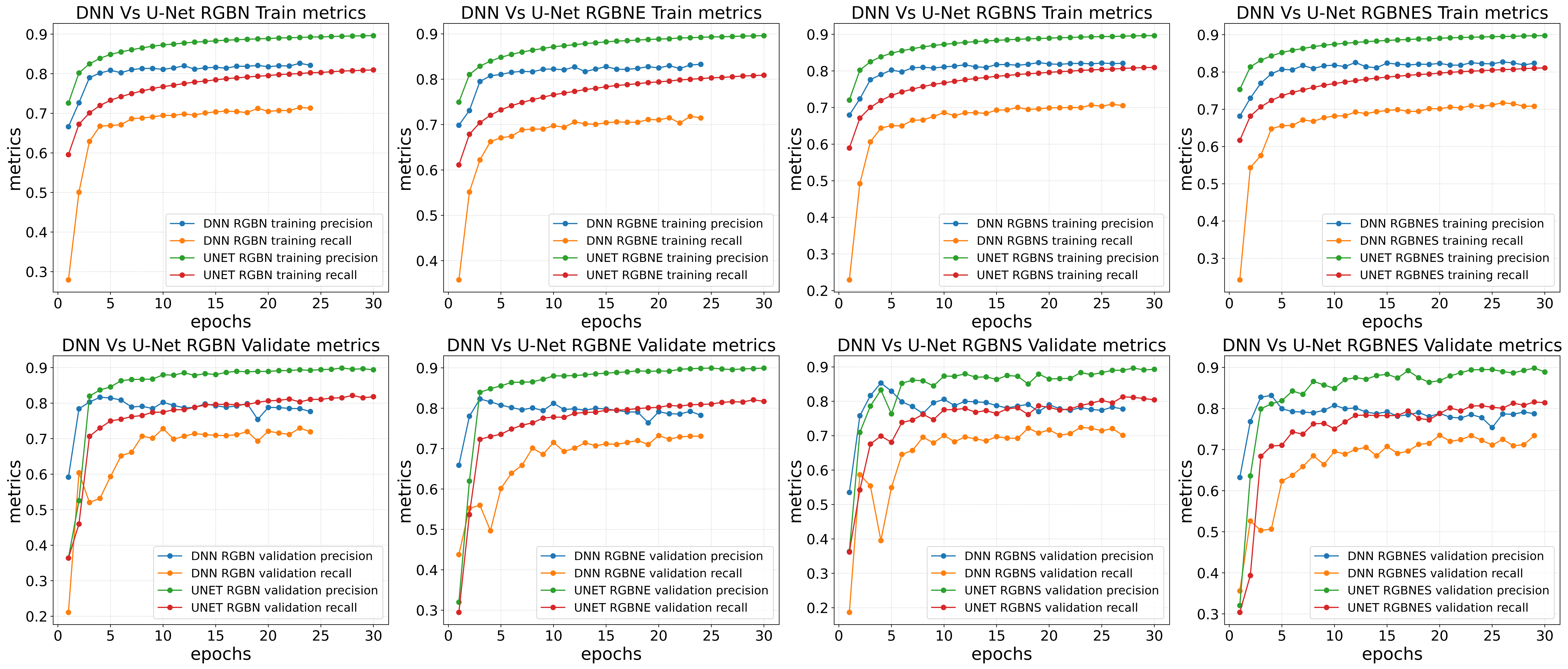} 
\caption[Plot of the comparison of the DNN model with the U-Net model during training and validation for precision and recall.]{Plot of the comparison of the DNN model with the U-Net model during training and validation for precision and recall.}
\label{fig:metrics_plot_dnn_unet_model_comparison}
\end{figure}

When comparing the test evaluation result between the two architectures, the U-Net model consistently performed better than the DNN model both in the training and validation across all metrics (categorical accuracy, precision, recall, and F1 score) as displayed in the training and validation plots in figure~\ref{fig:metrics_plot_dnn_unet_model_comparison}. As displayed in the validation plots of figure ~\ref{fig:metrics_plot_dnn_unet_model_comparison}, the DNN model initially starts out by performing better on the lower epochs (> 5), but as the epochs increases U-Net performance is much better than the DNN model. Similar results can be seen with the independent validation and area calculations analyses as explained above. As the U-Net approach has been shown to be the preferred method, this signifies the importance of employing a DL architecture that incorporates a patch-based neighborhood of the land cover in the modeling stage. This is most likely due to the U-Net input training data accessing spatially larger data patches compared to the DNN model (pixel). In addition, since image augmentation including flipping, random brightness and random contrast were applied to the U-Net model, it had access to more variations in the data set as compared to the DNN model. And finally the unique spatial context of rice fields observed as both on valley floors (in contiguous dense fields) as well as terraced (often in linear configurations) is captured in a patch-based training sampling approach. This additional patch context and spatial variation is likely a key contribute to the U-Net model sets overall higher performance compared to the DNN. Therefore, it is recommended for practitioners to keep attention to spatial constraints of the focal phenomenon/class and select model architectures and approaches that sufficiently capture this.

This study relied on only core bands from Planet, i.e. (R, G, B, N) and excluded the construction of additional derived indices such as NDVI, EVI, NDWI, SAVI, MSAVI, MTVI, VARI, and TGI. From the preliminary testing, it was observed that with the addition of optical features only offered minor improvement over the  test data set. However, all the combinations of those features for the DL models were not exhausted. Nonetheless, unlike tree based model such as Random Forest which benefit from employing extensive feature engineering approaches in the form of constructing indices \cite{belgiu2016random}, DL based models are able to capture non-linear relationships using only the R, G, B, N bands due to the multi-layer learning architecture \cite{bengio2013representation, lecun2015deep, zhang2016deep, yuan2020deep}. The additional benefit of not including additional features is that it enables faster model training and thus inferences.

The final rice maps were produced at 10-m resolution. As future work and areas to explore our team in conjunction with partners are interested employing U-Net modeling approaches in association with very high spatial and temporal resolution sensors. Practitioners in Bhutan have expressed interest to explore and leverage even higher resolution drones based sensor platforms ~\cite{Flying_Labs_Blog_2023}. The drone platforms are capable of capturing very high resolution (up to few cm) with high temporal accuracy, which widens the opportunities for near real time high resolution crop maps that are not hindered by clouds or other atmospheric conditions.

A limitation of the current modelling approach is that the model was trained using a Remote Sensing composite before growing season and during growing season (between June to September). As such, limited phenological characteristics were included in the model. Methods that can incorporate phenology based information can be used. Recurrent Neural Networks are often used to map phenological characteristics. For example, \cite{ndikumana2018deep} used a Long Short Term Memory (LSTM) \cite{hochreiter1996lstm} and Gated Recurrent Unit (GRU) \cite{cho2014learning} to map 11 agricultural classes including rice, whereas Sun et al. (2019) \cite{sun2019county} implemented predicted Soybean Yield with a CNN-LSTM Model using both spatial and temporal information. Lately, the use of Transformers has been increasing in Remote Sensing applications \cite{aleissaee2023transformers}. Crop Mapping may greatly benefit from fine-tuning or transfer learning using transformer-based model that are pre-trained on remote sensing pixel-time series data like PRESTO \cite{tseng2023lightweight} or PRITHVI \cite{jakubik2023foundation}.

Similarly, while depth-wise separable convolutions were used, the U-Net architecture employed was standard. Multiple studies have used variations on the encoding, for e.g. ~\cite{mayer2021deep} uses Visual Geometry Group (VGG) 19 model architecture ~\cite{simonyan2014very} for the encoder part and ~\cite{poortinga2021mapping} uses MobileNetV2 ~\cite{howard2017mobilenets} as the encoder block. The accuracy may be improved when using these customized variations. This study did not experimented with these approaches as it was beyond the scope of the analysis. However, it should also be noted that data scarcity is a challenge in this area, so future study may also benefit from exploring solution to data scarcity. Further, the S1 images were RTC corrected in the GEE were performed using the angular based method developed by Vollrath et al \cite{vollrath2020}. However, other RTC correction software, algorithms or data product were not employed which may provide improved result \cite{flores2023evaluating}.

\section{Conclusions}


From the study, the authors conclude that the DNN and U-Net DL approaches applied are able to map crop type and crop extent of rice and demonstrate that DL methods can be used in combination with the survey based approaches. Also it was identified that the U-Net patch-based neighbourhood algorithm performed better than point-based algorithm DNN through head to head model comparisons. From this study, it was concluded that  regional land cover products can be employed as a weak labels approach to capture different strata for addressing pronounced class imbalance challenges. Additionally, from the preliminary testing the efficacy of DL approaches to capture non-linear features in the modeling stage was demonstrated and underscores the limited need for excessive feature engineering required compared to classical ML approaches. And finally through the study the additional designed independent validation step is advised to thoroughly evaluate final model performance and recommend to future practitioners.

\section{Funding}

This research was funded through the US Agency for International Development (USAID) and NASA initiative Cooperative Agreement Number: AID486-A-14-00002. Individuals affiliated with the University of Alabama in Huntsville (UAH) are funded through the NASA Applied Sciences Capacity Building Program, NASA Cooperative Agreement: 80MSFC22M004.

\section{Declaration of competing interest}

The authors declare that they have no known competing financial interests or personal relationships that could have appeared to influence the work reported in this paper.

\section{Data Availability Statement}

All the source code and methodologies developed for this study is open and can be obtained from GitHub at \href{https://github.com/biplovbhandari/aces}{https://github.com/biplovbhandari/aces}. The Collect Earth Online (CEO) can be viewed at \href{https://app.collect.earth/collection?projectId=34728}{https://app.collect.earth/collection?projectId=34728} \cite{admin_openforis_org_2024_10552092}.

\section{Acknowledgement}

The authors would like to thank our partners at the Bhutan Department of Agriculture (DoA), Bhutan National Statistics Bureau (NBS), and Bhutan Ecological Society (BES) for their continuous engagement, support, and critical feedback. Additionally, the authors would like to thank the advancing Science, Technology, Engineering, and Mathematics (STEM) in Bhutan which was part of the Inter-Agency US Department of State and NASA Inter-Agency Agreement which spurred the initial research.


\renewcommand{\refname}{\spacedlowsmallcaps{References}} 

\bibliographystyle{unsrt}

\bibliography{article} 

\begin{thebibliography}{100}

\bibitem{food2017future}
Food and Agriculture~Organization of~the United~Nations.
\newblock {\em The future of food and agriculture: Trends and challenges}.
\newblock Fao, 2017.

\bibitem{liu2020agricultural}
Jianxu Liu, Mengjiao Wang, Li~Yang, Sanzidur Rahman, and Songsak Sriboonchitta.
\newblock Agricultural productivity growth and its determinants in south and
  southeast asian countries.
\newblock {\em Sustainability}, 12(12):4981, 2020.

\bibitem{weiss2020remote}
Marie Weiss, Fr{\'e}d{\'e}ric Jacob, and Grgory Duveiller.
\newblock Remote sensing for agricultural applications: A meta-review.
\newblock {\em Remote sensing of environment}, 236:111402, 2020.

\bibitem{tariq2023mapping}
Aqil Tariq, Jianguo Yan, Alexandre~S Gagnon, Mobushir Riaz~Khan, and Faisal
  Mumtaz.
\newblock Mapping of cropland, cropping patterns and crop types by combining
  optical remote sensing images with decision tree classifier and random
  forest.
\newblock {\em Geo-Spatial Information Science}, 26(3):302--320, 2023.

\bibitem{lobell2004cropland}
David~B Lobell and Gregory~P Asner.
\newblock Cropland distributions from temporal unmixing of modis data.
\newblock {\em Remote Sensing of Environment}, 93(3):412--422, 2004.

\bibitem{pan2012winter}
Yaozhong Pan, Le~Li, Jinshui Zhang, Shunlin Liang, Xiufang Zhu, and Damien
  Sulla-Menashe.
\newblock Winter wheat area estimation from modis-evi time series data using
  the crop proportion phenology index.
\newblock {\em Remote Sensing of Environment}, 119:232--242, 2012.

\bibitem{XIAO2005480}
Xiangming Xiao, Stephen Boles, Jiyuan Liu, Dafang Zhuang, Steve Frolking,
  Changsheng Li, William Salas, and Berrien Moore.
\newblock Mapping paddy rice agriculture in southern china using multi-temporal
  modis images.
\newblock {\em Remote Sensing of Environment}, 95(4):480--492, 2005.

\bibitem{XIAO200695}
Xiangming Xiao, Stephen Boles, Steve Frolking, Changsheng Li, Jagadeesh~Y.
  Babu, William Salas, and Berrien Moore.
\newblock Mapping paddy rice agriculture in south and southeast asia using
  multi-temporal modis images.
\newblock {\em Remote Sensing of Environment}, 100(1):95--113, 2006.

\bibitem{dong2016mapping}
Jinwei Dong, Xiangming Xiao, Michael~A Menarguez, Geli Zhang, Yuanwei Qin,
  David Thau, Chandrashekhar Biradar, and Berrien Moore~III.
\newblock Mapping paddy rice planting area in northeastern asia with landsat 8
  images, phenology-based algorithm and google earth engine.
\newblock {\em Remote sensing of environment}, 185:142--154, 2016.

\bibitem{mishra2023high}
Bhogendra Mishra, Rupesh Bhandari, Krishna~Prasad Bhandari, Dinesh~Mani
  Bhandari, Nirajan Luintel, A~Dahal, and Shobha Poudel.
\newblock High-resolution mapping of seasonal crop pattern using sentinel
  imagery in mountainous region of nepal: A semi-automatic approach.
\newblock {\em Geomatics}, 3(2):312--327, 2023.

\bibitem{park2018classification}
Seonyoung Park, Jungho Im, Seohui Park, Cheolhee Yoo, Hyangsun Han, and
  Jinyoung Rhee.
\newblock Classification and mapping of paddy rice by combining landsat and sar
  time series data.
\newblock {\em Remote Sensing}, 10(3):447, 2018.

\bibitem{ma2020unsupervised}
Zhe Ma, Zhe Liu, Yuanyuan Zhao, Lin Zhang, Diyou Liu, Tianwei Ren, Xiaodong
  Zhang, and Shaoming Li.
\newblock An unsupervised crop classification method based on principal
  components isometric binning.
\newblock {\em ISPRS International Journal of Geo-Information}, 9(11):648,
  2020.

\bibitem{mayaux2004new}
Philippe Mayaux, Etienne Bartholom{\'e}, Steffen Fritz, and Alan Belward.
\newblock A new land-cover map of africa for the year 2000.
\newblock {\em Journal of biogeography}, 31(6):861--877, 2004.

\bibitem{gao2006blending}
Feng Gao, Jeff Masek, Matt Schwaller, and Forrest Hall.
\newblock On the blending of the landsat and modis surface reflectance:
  Predicting daily landsat surface reflectance.
\newblock {\em IEEE Transactions on Geoscience and Remote sensing},
  44(8):2207--2218, 2006.

\bibitem{zhu2010enhanced}
Xiaolin Zhu, Jin Chen, Feng Gao, Xuehong Chen, and Jeffrey~G Masek.
\newblock An enhanced spatial and temporal adaptive reflectance fusion model
  for complex heterogeneous regions.
\newblock {\em Remote Sensing of Environment}, 114(11):2610--2623, 2010.

\bibitem{gevaert2015comparison}
Caroline~M Gevaert and F~Javier Garc{\'\i}a-Haro.
\newblock A comparison of starfm and an unmixing-based algorithm for landsat
  and modis data fusion.
\newblock {\em Remote sensing of Environment}, 156:34--44, 2015.

\bibitem{carrao2008contribution}
Hugo Carr{\~a}o, Paulo Gon{\c{c}}alves, and M{\'a}rio Caetano.
\newblock Contribution of multispectral and multitemporal information from
  modis images to land cover classification.
\newblock {\em Remote Sensing of Environment}, 112(3):986--997, 2008.

\bibitem{zhang2009mapping}
Yuan Zhang, Cuizhen Wang, Jiaping Wu, Jiaguo Qi, and William~A Salas.
\newblock Mapping paddy rice with multitemporal alos/palsar imagery in
  southeast china.
\newblock {\em International journal of Remote sensing}, 30(23):6301--6315,
  2009.

\bibitem{gebhardt2014mad}
Steffen Gebhardt, Thilo Wehrmann, Miguel Angel~Mu{\~n}oz Ruiz, Pedro Maeda,
  Jesse Bishop, Matthias Schramm, Rene Kopeinig, Oliver Cartus, Josef
  Kellndorfer, Rainer Ressl, et~al.
\newblock Mad-mex: Automatic wall-to-wall land cover monitoring for the mexican
  redd-mrv program using all landsat data.
\newblock {\em Remote Sensing}, 6(5):3923--3943, 2014.

\bibitem{mayer2023employing}
Timothy Mayer, Biplov Bhandari, Filoteo~G{\'o}mez Mart{\'\i}nez, Kaitlin
  Walker, Stephanie~A Jim{\'e}nez, Meryl Kruskopf, Micky Maganini, Aparna
  Phalke, Tshering Wangchen, Loday Phuntsho, et~al.
\newblock Employing the agricultural classification and estimation service
  (aces) for mapping smallholder rice farms in bhutan.
\newblock {\em Frontiers in Environmental Science}, 11:1137835, 2023.

\bibitem{o2020improved}
Kristen O’Shea, Jillian LaRoe, Anthony Vorster, Nicholas Young, Paul
  Evangelista, Timothy Mayer, Daniel Carver, Eli Simonson, Vanesa Martin, Paul
  Radomski, et~al.
\newblock Improved remote sensing methods to detect northern wild rice (zizania
  palustris l.).
\newblock {\em Remote Sensing}, 12(18):3023, 2020.

\bibitem{yu2023ricemapengine}
Zhiqi Yu, Liping Di, Sravan Shrestha, Chen Zhang, Liying Guo, Faisal Qamar, and
  Timothy~J Mayer.
\newblock Ricemapengine: A google earth engine-based web application for fast
  paddy rice mapping.
\newblock {\em IEEE Journal of Selected Topics in Applied Earth Observations
  and Remote Sensing}, 2023.

\bibitem{clark2012land}
Matthew~L Clark, T~Mitchell Aide, and George Riner.
\newblock Land change for all municipalities in latin america and the caribbean
  assessed from 250-m modis imagery (2001--2010).
\newblock {\em Remote Sensing of Environment}, 126:84--103, 2012.

\bibitem{singha2019high}
Mrinal Singha, Jinwei Dong, Geli Zhang, and Xiangming Xiao.
\newblock High resolution paddy rice maps in cloud-prone bangladesh and
  northeast india using sentinel-1 data.
\newblock {\em Scientific data}, 6(1):1--10, 2019.

\bibitem{lasko2018mapping}
Kristofer Lasko, Krishna~Prasad Vadrevu, Vinh~Tuan Tran, and Christopher
  Justice.
\newblock Mapping double and single crop paddy rice with sentinel-1a at varying
  spatial scales and polarizations in hanoi, vietnam.
\newblock {\em IEEE Journal of Selected Topics in Applied Earth Observations
  and Remote Sensing}, 11(2):498--512, 2018.

\bibitem{poortinga2021mapping}
Ate Poortinga, Nyein~Soe Thwal, Nishanta Khanal, Timothy Mayer, Biplov
  Bhandari, Kel Markert, Andrea~P Nicolau, John Dilger, Karis Tenneson,
  Nicholas Clinton, et~al.
\newblock Mapping sugarcane in thailand using transfer learning, a lightweight
  convolutional neural network, nicfi high resolution satellite imagery and
  google earth engine.
\newblock {\em ISPRS Open Journal of Photogrammetry and Remote Sensing},
  1:100003, 2021.

\bibitem{mayer2021deep}
Timothy Mayer, Ate Poortinga, Biplov Bhandari, Andrea~P Nicolau, Kel Markert,
  Nyein~Soe Thwal, Amanda Markert, Arjen Haag, John Kilbride, Farrukh Chishtie,
  et~al.
\newblock Deep learning approach for sentinel-1 surface water mapping
  leveraging google earth engine.
\newblock {\em ISPRS Open Journal of Photogrammetry and Remote Sensing},
  2:100005, 2021.

\bibitem{parekh2021automatic}
Jash~R Parekh, Ate Poortinga, Biplov Bhandari, Timothy Mayer, David Saah, and
  Farrukh Chishtie.
\newblock Automatic detection of impervious surfaces from remotely sensed data
  using deep learning.
\newblock {\em Remote Sensing}, 13(16):3166, 2021.

\bibitem{lv2020delineation}
Yahui Lv, Chao Zhang, Wenju Yun, Lulu Gao, Huan Wang, Jiani Ma, Hongju Li, and
  Dehai Zhu.
\newblock The delineation and grading of actual crop production units in modern
  smallholder areas using rs data and mask r-cnn.
\newblock {\em Remote Sensing}, 12(7):1074, 2020.

\bibitem{gomez2016optical}
Cristina G{\'o}mez, Joanne~C White, and Michael~A Wulder.
\newblock Optical remotely sensed time series data for land cover
  classification: A review.
\newblock {\em ISPRS Journal of photogrammetry and Remote Sensing}, 116:55--72,
  2016.

\bibitem{tashi2018mapping}
Dorji Tashi.
\newblock {\em Mapping change in rice cultivation using geospatial science in
  the Paro valley, Bhutan from 1995-2011}.
\newblock PhD thesis, Flinders University, College of Science and Engineering.,
  2018.

\bibitem{tshewang2016weed}
Sangay Tshewang, Brian~M Sindel, Mahesh Ghimiray, and Bhagirath~Singh Chauhan.
\newblock Weed management challenges in rice (oryza sativa l.) for food
  security in bhutan: A review.
\newblock {\em Crop Protection}, 90:117--124, 2016.

\bibitem{WB_Bhutan}
World Bank.
\newblock Bhutan overview: Development news, research, data | world bank.

\bibitem{world2019bhutan}
World Bank.
\newblock Bhutan urban policy notes: Regional development and economic
  transformation.
\newblock 2019.

\bibitem{FRIEDL2010168}
Mark~A. Friedl, Damien Sulla-Menashe, Bin Tan, Annemarie Schneider, Navin
  Ramankutty, Adam Sibley, and Xiaoman Huang.
\newblock Modis collection 5 global land cover: Algorithm refinements and
  characterization of new datasets.
\newblock {\em Remote Sensing of Environment}, 114(1):168--182, 2010.

\bibitem{buchhorn2020copernicus}
Marcel Buchhorn, Myroslava Lesiv, Nandin-Erdene Tsendbazar, Martin Herold, Luc
  Bertels, and Bruno Smets.
\newblock Copernicus global land cover layers—collection 2.
\newblock {\em Remote Sensing}, 12(6):1044, 2020.

\bibitem{brown2022dynamic}
Christopher~F Brown, Steven~P Brumby, Brookie Guzder-Williams, Tanya Birch,
  Samantha~Brooks Hyde, Joseph Mazzariello, Wanda Czerwinski, Valerie~J
  Pasquarella, Robert Haertel, Simon Ilyushchenko, et~al.
\newblock Dynamic world, near real-time global 10 m land use land cover
  mapping.
\newblock {\em Scientific Data}, 9(1):251, 2022.

\bibitem{SAAH2020101979}
David Saah, Karis Tenneson, Ate Poortinga, Quyen Nguyen, Farrukh Chishtie,
  Khun~San Aung, Kel~N. Markert, Nicholas Clinton, Eric~R. Anderson, Peter
  Cutter, Joshua Goldstein, Ian~W. Housman, Biplov Bhandari, Peter~V. Potapov,
  Mir Matin, Kabir Uddin, Hai~N. Pham, Nishanta Khanal, Sajana Maharjan,
  Walter~L. Ellenberg, Birendra Bajracharya, Radhika Bhargava, Paul Maus,
  Matthew Patterson, Africa~Ixmucane Flores-Anderson, Jeffrey Silverman,
  Chansopheaktra Sovann, Phuong~M. Do, Giang~V. Nguyen, Soukanh Bounthabandit,
  Raja~Ram Aryal, Su~Mon Myat, Kei Sato, Erik Lindquist, Marija Kono, Jeremy
  Broadhead, Peeranan Towashiraporn, and David Ganz.
\newblock Primitives as building blocks for constructing land cover maps.
\newblock {\em International Journal of Applied Earth Observation and
  Geoinformation}, 85:101979, 2020.

\bibitem{uddin2021regional}
Kabir Uddin, Mir~A Matin, Nishanta Khanal, Sajana Maharjan, Birendra
  Bajracharya, Karis Tenneson, Ate Poortinga, Nguyen~Hanh Quyen, Raja~Ram
  Aryal, David Saah, et~al.
\newblock Regional land cover monitoring system for hindu kush himalaya.
\newblock In {\em Earth observation science and applications for risk reduction
  and enhanced resilience in Hindu Kush Himalaya region}, pages 103--125.
  Springer, 2021.

\bibitem{gumma2011mapping}
Murali~Krishna Gumma, Andrew Nelson, Prasad~S Thenkabail, and Amrendra~N Singh.
\newblock Mapping rice areas of south asia using modis multitemporal data.
\newblock {\em Journal of applied remote sensing}, 5(1):053547--053547, 2011.

\bibitem{WB_2017}
The~World Bank.
\newblock Food security and agriculture productivity project, 4 2017.

\bibitem{gurung2023identification}
Lily Gurung, Manoj Chhetri, and Parshu~Ram Dhungyel.
\newblock Identification of appropriate drone technology and its application
  areas for agriculture automation in bhutan.
\newblock {\em Bhutan Journal of Research and Development}, (2), Feb. 2023.

\bibitem{dorji2023agricultural}
Kinley Dorji, Judith Miller, and Shubiao Wu.
\newblock Agricultural interventions in the bhutanese context for
  sustainability—a documentary analysis using a thematic conceptual
  framework.
\newblock {\em Sustainability}, 15(5):4177, 2023.

\bibitem{mccartney2021advancing}
Sean McCartney, Amanda Clayton, Kenton Ross, Lauren Childs-Gleason, Karen
  Allsbrook, Stephanie Burke, Michael Ruiz, Amber~Rochelle Williams, Hayley
  Pippin, Celeste Gambino, et~al.
\newblock Advancing stem in bhutan through increased earth observation
  capacity: Applying nasa develop’s model to bhutanese scholars.
\newblock In {\em AGU Fall Meeting 2021}, number SY55B-0354, 2021.

\bibitem{breiman2001random}
Leo Breiman.
\newblock Random forests.
\newblock {\em Machine learning}, 45(1):5--32, 2001.

\bibitem{gorelick2017google}
Noel Gorelick, Matt Hancher, Mike Dixon, Simon Ilyushchenko, David Thau, and
  Rebecca Moore.
\newblock Google earth engine: Planetary-scale geospatial analysis for
  everyone.
\newblock {\em Remote sensing of Environment}, 202:18--27, 2017.

\bibitem{NSB_Ag_2019}
Renewable Natural Resources Statistics~Division (RSD).
\newblock Agriculture statistics 2020, 2020.

\bibitem{NSB_Ag_2020}
Renewable Natural Resources Statistics~Division (RSD).
\newblock Agriculture statistics 2020, 2021.

\bibitem{NSB_Ag_2021}
National~Statistics Bureau.
\newblock Agriculture survey report 2021, 2022.

\bibitem{Figueredo:2009dg}
A.~J. Figueredo and P.~S.~A. Wolf.
\newblock Assortative pairing and life history strategy - a cross-cultural
  study.
\newblock {\em Human Nature}, 20:317--330, 2009.

\bibitem{gilani2015decadal}
Hammad Gilani, Him~Lal Shrestha, MSR Murthy, Phuntso Phuntso, Sudip Pradhan,
  Birendra Bajracharya, and Basanta Shrestha.
\newblock Decadal land cover change dynamics in bhutan.
\newblock {\em Journal of environmental management}, 148:91--100, 2015.

\bibitem{NSB2023}
National~Statistics Bureau.
\newblock Statistical yearbook – national statistics bureau, 10 2023.

\bibitem{DoSAM2023}
DoSAM~Department of~Surveying and Mapping.
\newblock Land use land cover 2020, 2023.

\bibitem{gyeltshen2019integrate}
Kinzang Gyeltshen and Sheetal Sharma.
\newblock Integrated plant nutrition system modules for major crops and
  cropping systems in south asia.
\newblock 2019.

\bibitem{Agriculture2017}
International~Center for Tropical~Agriculture and World Bank.
\newblock Climate-smart agriculture in bhutan, 9 2017.

\bibitem{chhogyel2018climate}
Ngawang Chhogyel and Lalit Kumar.
\newblock Climate change and potential impacts on agriculture in bhutan: a
  discussion of pertinent issues.
\newblock {\em Agriculture \& food security}, 7(1):1--13, 2018.

\bibitem{Namgay2021}
Yenten Namgay, Tirtha~Bdr. Katwal, Yadunath Bajgai, and Padam~Lal Giri.
\newblock Agronomic parameters of high-altitude rice varieties and their
  relation to temperature at different growth stages.
\newblock {\em Bhutanese Journal of Agriculture}, 4:63--76, 2 2021.

\bibitem{jena2012advances}
KK~Jena.
\newblock {\em Advances in temperate rice research}.
\newblock Int. Rice Res. Inst., 2012.

\bibitem{planet_nicfi_data}
Planet Labs.
\newblock Nicfi satellite data program, 2022.

\bibitem{lemajic2018new}
Blanka Vajsova~Slavko Lemajic and P{\"a}r~Johan {\AA}strand.
\newblock New sensors benchmark report on planetscope.
\newblock 2018.

\bibitem{planet_indices}
Planet Labs.
\newblock Planet remote sensing indices.

\bibitem{snap_toolbox}
European Space~Agency (ESA).
\newblock Sentinel-1 toolbox - sentinel online.

\bibitem{farr2007shuttle}
Tom~G Farr, Paul~A Rosen, Edward Caro, Robert Crippen, Riley Duren, Scott
  Hensley, Michael Kobrick, Mimi Paller, Ernesto Rodriguez, Ladislav Roth,
  et~al.
\newblock The shuttle radar topography mission.
\newblock {\em Reviews of geophysics}, 45(2), 2007.

\bibitem{markert2020comparing}
Kel~N Markert, Amanda~M Markert, Timothy Mayer, Claire Nauman, Arjen Haag, Ate
  Poortinga, Biplov Bhandari, Nyein~Soe Thwal, Thannarot Kunlamai, Farrukh
  Chishtie, et~al.
\newblock Comparing sentinel-1 surface water mapping algorithms and radiometric
  terrain correction processing in southeast asia utilizing google earth
  engine.
\newblock {\em Remote Sensing}, 12(15):2469, 2020.

\bibitem{small2011flattening}
David Small.
\newblock Flattening gamma: Radiometric terrain correction for sar imagery.
\newblock {\em IEEE Transactions on Geoscience and Remote Sensing},
  49(8):3081--3093, 2011.

\bibitem{vollrath2020}
Andreas Vollrath, Adugna Mullissa, and Johannes Reiche.
\newblock Angular-based radiometric slope correction for sentinel-1 on google
  earth engine.
\newblock {\em Remote Sensing}, 12(11), 2020.

\bibitem{lee2008improved}
Jong-Sen Lee, Jen-Hung Wen, Thomas~L Ainsworth, Kun-Shan Chen, and Abel~J Chen.
\newblock Improved sigma filter for speckle filtering of sar imagery.
\newblock {\em IEEE Transactions on Geoscience and Remote Sensing},
  47(1):202--213, 2008.

\bibitem{tachikawa2011characteristics}
Tetsushi Tachikawa, Masami Hato, Manabu Kaku, and Akira Iwasaki.
\newblock Characteristics of aster gdem version 2.
\newblock In {\em 2011 IEEE international geoscience and remote sensing
  symposium}, pages 3657--3660. IEEE, 2011.

\bibitem{danielson2011global}
Jeffrey~J Danielson, Dean~B Gesch, et~al.
\newblock Global multi-resolution terrain elevation data 2010 (gmted2010).
\newblock 2011.

\bibitem{gesch2002national}
Dean Gesch, Michael Oimoen, Susan Greenlee, Charles Nelson, Michael Steuck, and
  Dean Tyler.
\newblock The national elevation dataset.
\newblock {\em Photogrammetric engineering and remote sensing}, 68(1):5--32,
  2002.

\bibitem{haris_iqbal_2018_2526396}
Haris Iqbal.
\newblock Harisiqbal88/plotneuralnet v1.0.0, December 2018.

\bibitem{chollet2015keras}
Fran\c{c}ois Chollet et~al.
\newblock Keras.
\newblock \url{https://keras.io}, 2015.

\bibitem{ronneberger2015u}
Olaf Ronneberger, Philipp Fischer, and Thomas Brox.
\newblock U-net: Convolutional networks for biomedical image segmentation.
\newblock In {\em Medical Image Computing and Computer-Assisted
  Intervention--MICCAI 2015: 18th International Conference, Munich, Germany,
  October 5-9, 2015, Proceedings, Part III 18}, pages 234--241. Springer, 2015.

\bibitem{sandler2018mobilenetv2}
Mark Sandler, Andrew Howard, Menglong Zhu, Andrey Zhmoginov, and Liang-Chieh
  Chen.
\newblock Mobilenetv2: Inverted residuals and linear bottlenecks.
\newblock In {\em Proceedings of the IEEE conference on computer vision and
  pattern recognition}, pages 4510--4520, 2018.

\bibitem{kingma2017adam}
Diederik~P. Kingma and Jimmy Ba.
\newblock Adam: A method for stochastic optimization, 2017.

\bibitem{shorten2019survey}
Connor Shorten and Taghi~M Khoshgoftaar.
\newblock A survey on image data augmentation for deep learning.
\newblock {\em Journal of big data}, 6(1):1--48, 2019.

\bibitem{he2016deep}
Kaiming He, Xiangyu Zhang, Shaoqing Ren, and Jian Sun.
\newblock Deep residual learning for image recognition.
\newblock In {\em Proceedings of the IEEE conference on computer vision and
  pattern recognition}, pages 770--778, 2016.

\bibitem{hartigan1979algorithm}
John~A Hartigan and Manchek~A Wong.
\newblock Algorithm as 136: A k-means clustering algorithm.
\newblock {\em Journal of the royal statistical society. series c (applied
  statistics)}, 28(1):100--108, 1979.

\bibitem{lloyd1982least}
Stuart Lloyd.
\newblock Least squares quantization in pcm.
\newblock {\em IEEE transactions on information theory}, 28(2):129--137, 1982.

\bibitem{MSBuilding2023}
Microsoft.
\newblock Worldwide building footprints derived from satellite imagery, 2023.

\bibitem{soares2020landslide}
Lucas~P Soares, Helen~C Dias, and Carlos~H Grohmann.
\newblock Landslide segmentation with u-net: Evaluating different sampling
  methods and patch sizes.
\newblock {\em arXiv preprint arXiv:2007.06672}, 2020.

\bibitem{van1979information}
C~Van~Rijsbergen.
\newblock Information retrieval: theory and practice.
\newblock In {\em Proceedings of the Joint IBM/University of Newcastle upon
  Tyne Seminar on Data Base Systems}, volume~79, 1979.

\bibitem{chicco2020advantages}
Davide Chicco and Giuseppe Jurman.
\newblock The advantages of the matthews correlation coefficient (mcc) over f1
  score and accuracy in binary classification evaluation.
\newblock {\em BMC genomics}, 21(1):1--13, 2020.

\bibitem{saah2019collect}
David Saah, Gary Johnson, Billy Ashmall, Githika Tondapu, Karis Tenneson, Matt
  Patterson, Ate Poortinga, Kel Markert, Nguyen~Hanh Quyen, Khun San~Aung,
  et~al.
\newblock Collect earth: An online tool for systematic reference data
  collection in land cover and use applications.
\newblock {\em Environmental Modelling \& Software}, 118:166--171, 2019.

\bibitem{tensorflow2015-whitepaper}
Mart\'{i}n Abadi, Ashish Agarwal, Paul Barham, Eugene Brevdo, Zhifeng Chen,
  Craig Citro, Greg~S. Corrado, Andy Davis, Jeffrey Dean, Matthieu Devin,
  Sanjay Ghemawat, Ian Goodfellow, Andrew Harp, Geoffrey Irving, Michael Isard,
  Yangqing Jia, Rafal Jozefowicz, Lukasz Kaiser, Manjunath Kudlur, Josh
  Levenberg, Dandelion Man\'{e}, Rajat Monga, Sherry Moore, Derek Murray, Chris
  Olah, Mike Schuster, Jonathon Shlens, Benoit Steiner, Ilya Sutskever, Kunal
  Talwar, Paul Tucker, Vincent Vanhoucke, Vijay Vasudevan, Fernanda Vi\'{e}gas,
  Oriol Vinyals, Pete Warden, Martin Wattenberg, Martin Wicke, Yuan Yu, and
  Xiaoqiang Zheng.
\newblock {TensorFlow}: Large-scale machine learning on heterogeneous systems,
  2015.
\newblock Software available from tensorflow.org.

\bibitem{belgiu2016random}
Mariana Belgiu and Lucian Dr{\u{a}}gu{\c{t}}.
\newblock Random forest in remote sensing: A review of applications and future
  directions.
\newblock {\em ISPRS journal of photogrammetry and remote sensing}, 114:24--31,
  2016.

\bibitem{bengio2013representation}
Yoshua Bengio, Aaron Courville, and Pascal Vincent.
\newblock Representation learning: A review and new perspectives.
\newblock {\em IEEE transactions on pattern analysis and machine intelligence},
  35(8):1798--1828, 2013.

\bibitem{lecun2015deep}
Yann LeCun, Yoshua Bengio, and Geoffrey Hinton.
\newblock Deep learning.
\newblock {\em nature}, 521(7553):436--444, 2015.

\bibitem{zhang2016deep}
Liangpei Zhang, Lefei Zhang, and Bo~Du.
\newblock Deep learning for remote sensing data: A technical tutorial on the
  state of the art.
\newblock {\em IEEE Geoscience and remote sensing magazine}, 4(2):22--40, 2016.

\bibitem{yuan2020deep}
Qiangqiang Yuan, Huanfeng Shen, Tongwen Li, Zhiwei Li, Shuwen Li, Yun Jiang,
  Hongzhang Xu, Weiwei Tan, Qianqian Yang, Jiwen Wang, et~al.
\newblock Deep learning in environmental remote sensing: Achievements and
  challenges.
\newblock {\em Remote Sensing of Environment}, 241:111716, 2020.

\bibitem{Flying_Labs_Blog_2023}
Jul 2023.

\bibitem{ndikumana2018deep}
Emile Ndikumana, Dinh Ho~Tong~Minh, Nicolas Baghdadi, Dominique Courault, and
  Laure Hossard.
\newblock Deep recurrent neural network for agricultural classification using
  multitemporal sar sentinel-1 for camargue, france.
\newblock {\em Remote Sensing}, 10(8):1217, 2018.

\bibitem{hochreiter1996lstm}
Sepp Hochreiter and J{\"u}rgen Schmidhuber.
\newblock Lstm can solve hard long time lag problems.
\newblock {\em Advances in neural information processing systems}, 9, 1996.

\bibitem{cho2014learning}
Kyunghyun Cho, Bart Van~Merri{\"e}nboer, Caglar Gulcehre, Dzmitry Bahdanau,
  Fethi Bougares, Holger Schwenk, and Yoshua Bengio.
\newblock Learning phrase representations using rnn encoder-decoder for
  statistical machine translation.
\newblock {\em arXiv preprint arXiv:1406.1078}, 2014.

\bibitem{sun2019county}
Jie Sun, Liping Di, Ziheng Sun, Yonglin Shen, and Zulong Lai.
\newblock County-level soybean yield prediction using deep cnn-lstm model.
\newblock {\em Sensors}, 19(20):4363, 2019.

\bibitem{aleissaee2023transformers}
Abdulaziz~Amer Aleissaee, Amandeep Kumar, Rao~Muhammad Anwer, Salman Khan,
  Hisham Cholakkal, Gui-Song Xia, and Fahad~Shahbaz Khan.
\newblock Transformers in remote sensing: A survey.
\newblock {\em Remote Sensing}, 15(7):1860, 2023.

\bibitem{tseng2023lightweight}
Gabriel Tseng, Ivan Zvonkov, Mirali Purohit, David Rolnick, and Hannah Kerner.
\newblock Lightweight, pre-trained transformers for remote sensing timeseries.
\newblock {\em arXiv preprint arXiv:2304.14065}, 2023.

\bibitem{jakubik2023foundation}
Johannes Jakubik, Sujit Roy, CE~Phillips, Paolo Fraccaro, Denys Godwin, Bianca
  Zadrozny, Daniela Szwarcman, Carlos Gomes, Gabby Nyirjesy, Blair Edwards,
  et~al.
\newblock Foundation models for generalist geospatial artificial intelligence.
\newblock {\em arXiv preprint arXiv:2310.18660}, 2023.

\bibitem{simonyan2014very}
Karen Simonyan and Andrew Zisserman.
\newblock Very deep convolutional networks for large-scale image recognition.
\newblock {\em arXiv preprint arXiv:1409.1556}, 2014.

\bibitem{howard2017mobilenets}
Andrew~G Howard, Menglong Zhu, Bo~Chen, Dmitry Kalenichenko, Weijun Wang,
  Tobias Weyand, Marco Andreetto, and Hartwig Adam.
\newblock Mobilenets: Efficient convolutional neural networks for mobile vision
  applications.
\newblock {\em arXiv preprint arXiv:1704.04861}, 2017.

\bibitem{flores2023evaluating}
Africa~I Flores-Anderson, Helen~Blue Parache, Vanesa Martin-Arias, Stephanie~A
  Jim{\'e}nez, Kelsey Herndon, Stefanie Mehlich, Franz~J Meyer, Shobhit
  Agarwal, Simon Ilyushchenko, Manoj Agarwal, et~al.
\newblock Evaluating sar radiometric terrain correction products:
  Analysis-ready data for users.
\newblock {\em Remote Sensing}, 15(21):5110, 2023.

\bibitem{admin_openforis_org_2024_10552092}
admin@openforis.org.
\newblock Bhutan aces 2.0\_paro data collection round 3, February 2024.

\end{thebibliography}


\end{document}